\documentclass[runningheads]{llncs}
\usepackage{graphicx}
\usepackage{comment}
\usepackage{amsmath,amssymb} % define this before the line numbering.
\usepackage{color}
\usepackage{epsfig}
\usepackage{float}
\usepackage{wrapfig}
\usepackage{booktabs}
\usepackage{multirow}
\usepackage{ifthen}
\usepackage{alltt}
\usepackage{newlfont} % for Box
\usepackage{enumitem}
\usepackage{textcomp}

\newcommand{\IEncoder}{E_I}
\newcommand{\PEncoder}{E_P}

\usepackage{array}
\newcolumntype{L}[1]{>{\raggedright\let\newline\\\arraybackslash\hspace{0pt}}m{#1}}
\newcolumntype{C}[1]{>{\centering\let\newline\\\arraybackslash\hspace{0pt}}m{#1}}
\newcolumntype{R}[1]{>{\raggedleft\let\newline\\\arraybackslash\hspace{0pt}}m{#1}}

\makeatletter
\newcommand{\printfnsymbol}[1]{%
  \textsuperscript{\@fnsymbol{#1}}%
}

\begin{document}
% \linenumbers
\pagestyle{headings}
\mainmatter

\title{One-Shot Identity-Preserving Portrait Reenactment} % Replace with your title

%\begin{comment}
% \titlerunning{ECCV-20 submission ID \ECCVSubNumber} 
% \authorrunning{ECCV-20 submission ID \ECCVSubNumber} 
% \author{Anonymous ECCV submission}
% \institute{Paper ID \ECCVSubNumber}
%\end{comment}
%******************

%\begin{comment}
\titlerunning{One-Shot Identity-Preserving Portrait Reenactment}
\author{Sitao Xiang \inst{1} \and
Yuming Gu \inst{1} \thanks{indicates joint second authors.} \and
Pengda Xiang \inst{1} \printfnsymbol{1} \and
Mingming He \inst{1} \and
Koki Nagano \inst{1} \and
Haiwei Chen \inst{1} \and
Hao Li \inst{1,2}
}
\authorrunning{Xiang et al.}
\institute{University of Southern California / USC Institute for Creative Technologies \and
Pinscreen}%\\
% \email{{\small \{sitaoxia, yuminggu, pxiang\}@usc.edu}, {\small he@ict.usc.edu}, {\small koki.nagano0219@gmail.com}, {\small chw9308@hotmail.com}, {\small hao@hao-li.com}}}
%\end{comment}
%******************
%\maketitle

% \begin{figure}[ht]
% %	\footnotesize
% %	\centering
% 	\includegraphics[width=0.95\linewidth]{figure/teaser.jpg}
% 	\caption{\small{The results generated by our method using a variety of identities with different pose/expressions (first row) to control the two target portraits (first columns).}}
% 	\label{fig:teaser}
% \end{figure}

% \begin{center}
% \includegraphics[width=0.95\linewidth]{figure/teaser.pdf}
% % \includegraphics[width=\linewidth]{latex/figs/teaserFinal.pdf}\\
% \end{center}
% \begin{center}
% \caption{\small{he results generated by our method using landmarks from the source images to drive the target images, which show that both the pose/expression from the source images and the identity from the target images are well preserved.}}
% \label{fig:teaser}
% \end{center}

\maketitle
%\vspace{-2em}
%\vspace*{-8pt}
\begin{figure}
\begin{center}
\includegraphics[width=0.99\linewidth]{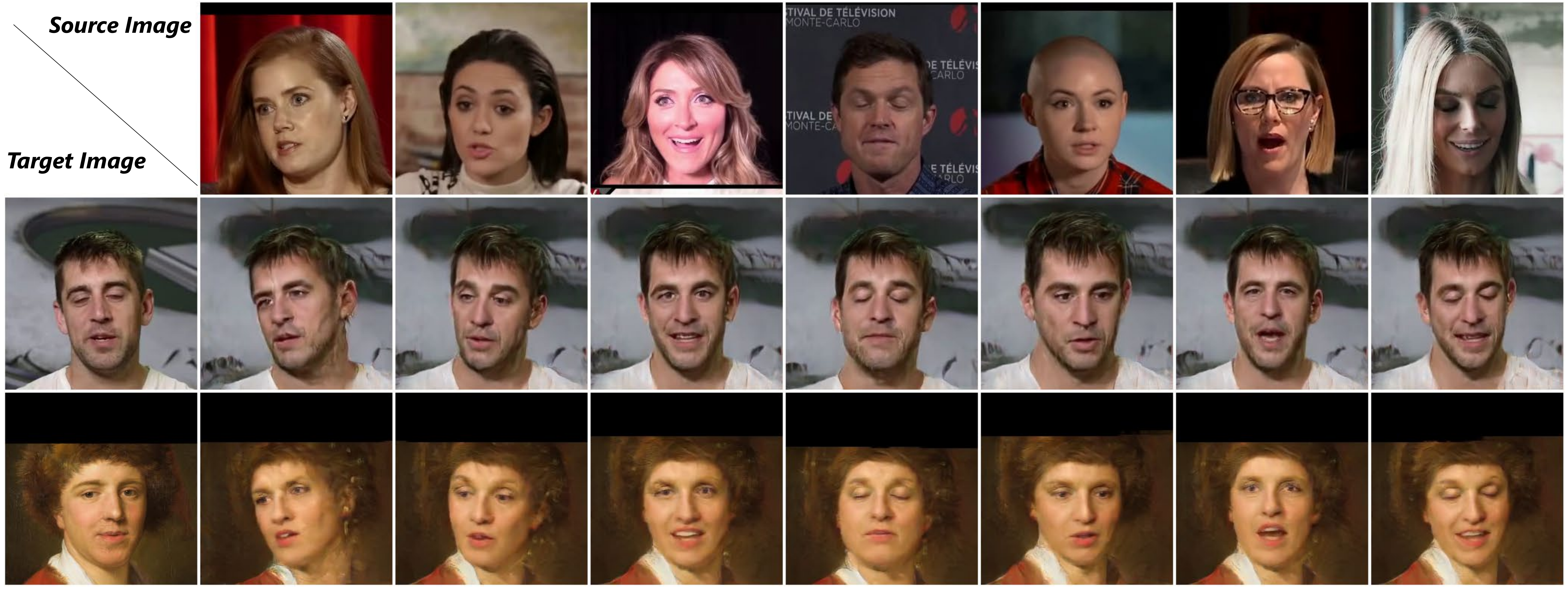}
% \includegraphics[width=\linewidth]{latex/figs/teaserFinal.pdf}\\=
% \begin{center}
\end{center}
\begin{center}
\vspace{-1.5em}
\caption{\small{Semantic image synthesis results produced by our method. Our method can not only synthesize images using only semantic segmentation masks as input, but also supports controllable synthesis via a reference style image.}}
\label{fig:teaser}
\end{center}
\vspace{-5em}
\end{figure}

% \begin{figure}[t]
% 	\footnotesize
% 	\centering
% 	\includegraphics[width=0.95\linewidth]{figure/teaser.pdf}
% 	\caption{The results generated by our method using landmarks from the source images to drive the target images, which show that both the pose/expression from the source images and the identity from the target images are well preserved.}
% 	\label{fig:pipeline}
% \end{figure}

\begin{abstract}
We present a deep learning-based framework for portrait reenactment from a single picture of a target (one-shot) and a video of a driving subject. Existing facial reenactment methods suffer from identity mismatch and produce inconsistent identities when a target and a driving subject are different (cross-subject), especially in one-shot settings. In this work, we aim to address identity preservation in cross-subject portrait reenactment from a single picture. We introduce a novel technique that can disentangle identity from expressions and poses, allowing identity preserving portrait reenactment even when the driver's identity is very different from that of the target. This is achieved by a novel landmark disentanglement network (LD-Net), which predicts personalized facial landmarks that combine the identity of the target with expressions and poses from a different subject. To handle portrait reenactment from unseen subjects, we also introduce a feature dictionary-based generative adversarial network (FD-GAN), which locally translates 2D landmarks into a personalized portrait, enabling one-shot portrait reenactment under large pose and expression variations. We validate the effectiveness of our identity disentangling capabilities via an extensive ablation study, and our method produces consistent identities for cross-subject portrait reenactment. Our comprehensive experiments show that our method significantly outperforms the state-of-the-art single-image facial reenactment methods. We will release our code and models for academic use.% and even few-shot learning methods which require subject-specific fine-tuning.
\end{abstract}

\section{Introduction}

Synthesizing facial animation from portrait images has a wide range of creative applications, including visual effects, multimedia messaging apps, and visual dubbing. Using someone's facial expressions and head pose in videos to drive the face of another person in a single image (known as portrait reenactment) is particularly popular due to its intuitive control and accessibility.

Traditionally, producing a realistically animated face is achieved by rendering a carefully digitized 3D head model with texture maps. Although many digital humans are still being produced this way for high-end visual effects and video games~\cite{Seymour:2017:MME},
this approach involves a tedious production effort including large teams of digital artists and often relies on complex 3D capture equipment~\cite{Ghosh:2011:MFC}. 
%and rendering technologies, creating a realistic digital human is an extremely labor-intensive task requiring weeks or months of effort by digital artists and remains a challenging task for a non-expert user. 

More recently, deep learning-based methods have gained significant attention due to their success in producing realistic face reenactment. 
%In particular, there has been active researches in generative adversarial networks (GAN)~\cite{goodfellow2014generative} including applications to human face synthesis and a significant body of work focus on conditional GANs~\cite{isola2016image} to constrain the synthesis for an image of interest. 
In particular, “DeepFakes” (\textit{e.g.}~\cite{deepfakes2019}) have become a widely used approach for end-to-end video-based face-swapping. However, it cannot generalize well to unseen subjects, requiring thousands of internet photos and often days of training for each subject. 

Currently, some advanced deep learning techniques for face image manipulation combine conditional GANs~\cite{isola2016image} with facial geometry information, such as 3D facial models~\cite{Nagano:2018:PRA,kim2018DeepVideo,Kim19NeuralDubbing} or 2D landmarks~\cite{zakharov2019few,nirkin2019fsgan}, to provide both better control and generalization capabilities w.r.t. arbitrary identities. These 3D model-based methods only work under specific conditions. They mostly rely on statistical face models and are often limited to certain face regions and linear shape variations. Furthermore, they lose accuracy for non-frontal portraits. However, our goal is to synthesize highly complex head poses, facial expressions and facial appearances (facial hair, stylized content, complex lighting conditions) as well as the image regions surrounding the face such as hair, background, etc. On the other hand, the 2D landmark-based methods are unable to properly preserve the accurate identity and complex facial expressions with the lack of appropriate landmark adaptation.

In this work, we wish to achieve a one-shot portrait reenactment of novel subjects (with no subject-specific training) for the cross-subject setting (meaning the ability to accommodate any driver).
In particular, we aim to address the problem of portrait reenactment from a single image of someone (the "target") using a sequence of 2D facial landmarks from a video of another person (the "source"). 

Our goal is to improve the preservation of identities within a cross-subject setting. Several challenges need to be addressed for identity-preserving face reenactment, particularly when a target and a source are different subjects. 
%\yajie{In particular, our goal is to achieve identity-preserving face reenactment in a cross-subject setting. There are a few technical challenges:} 
First, it is a non-trivial task to properly extract facial expressions/poses from person-specific facial features encoded using 2D facial landmark coordinates. \textit{For example, how can we determine whether a person has narrow eyes or is squinting and thus properly transfer identity-invariant motion to the target?} The second challenge lies in synthesizing photorealistic and recognizable results with arbitrary expression and novel views. The third challenge is to achieve the above from a single reference image of the target subject without relying on any person-specific training or fine-tuning~\cite{zakharov2019few}.  

We achieve this by introducing two novel sub-networks. The first, \textit{Landmark Disentanglement Network} (LD-Net), learns to disentangle the identity from the head poses and expressions, predicting facial landmarks that preserve the identity of the target while combining expressions and poses from another driving subject. The second, \textit{Feature Dictionary-based Generative Adversarial Network} (FD-GAN), learns to transform the landmark positions into a personalized video portrait of the subject depicted in a single target image, which allows subject-agnostic reenactment of a portrait that preserves the target's identity and can be applied to unseen subjects without subject-specific training. 
To summarize, we make the following technical contributions: (1) We introduce a novel one-shot learning method that enables portrait reenactment using the identity from a single image and expressions and poses from videos of another subject. (2) We present a novel network that disentangles the identity from 2D face landmarks for cross-subject portrait reenactment. (3) We also propose a feature dictionary-based generative network to synthesize a high-fidelity face image, which is applicable to new subjects. We evaluate each sub-network as well as the full method extensively via quantitative measurements and qualitative comparisons with the state-of-the-art methods, and demonstrate our method's ability to preserve the target subject's identity and to generalize to unseen subjects for cross-subject face reenactment.

\section{Related Work}

% \begin{enumerate}
%     \item \todo{Re-organize previous methods, 3D-based and 2D-based maybe. ( The 'coarse' groups' outline is questionable (for instance, "Facial Geometry and Appearance Capture" group might better be called something like "3D-based methods"))}
%     \item \todo{Describe flaws of previous methods. (The 'fine' paragraph organization seems to be random. The related methods are simply enumerated, being only weakly structured and not being assessed for flaws or disadvantages, or for relevance to the described method.)}
%     \item \todo{?Add related work Marionette~\cite{ha2019marionette}}
%     \item \todo{Add First-order-model~\cite{Siarohin_2019_NeurIPS}}
% \end{enumerate}{}

\paragraph{3D Graphics-based Methods.}
Research in 3D facial animation and rendering dates back several decades. %\cite{Egger20Years}. 
The seminal work of~\cite{BlanzVetter1999} showed that a morphable principal component model is effective in modeling human facial geometry and appearance. 
Over the years, a number of technical advances \cite{Egger20Years} have been made in capturing high-resolution textures and geometric details, as well as subject-specific expression deformations that are necessary to create realistic renderings of human face animations.
%However, the scans used to build the statistical model lack high-resolution texture and geometry details as well as person-specific expression deformations that are necessary to create a realistic rendering of human face animation. 
To capture facial textures, recent work uses deep learning-based analysis to infer photorealistic skin albedo maps from a single image~\cite{saito2016photorealistic,Gecer_2019_CVPR}. 
 %previous work inferred photorealistic skin albedo using a deep learning-based texture correlation analysis~\cite{saito2016photorealistic} and used a generative adversarial network (GAN) that learns a high-resolution UV space texture~\cite{Gecer_2019_CVPR}. 
 Olszewski et al.~\cite{olszewski2017realistic} showed that realistic face puppeteering is possible from a single picture by rendering a sequence of dynamic textures that are synthesized using generative adversarial networks (GANs).
% To realistically puppeteer a target face, Olszewski et al.~\cite{olszewski2017realistic} proposed a technique to transfer expression textures from a performer using GAN that operates in the UV texture space. 
 For face geometry, other recent approaches employ a non-linear morphable model~\cite{tran2019towards} to improve the fidelity, use a local regressor to enhance high-resolution details~\cite{cao2015real}, or use a pixel-to-pixel translation network to learn mesoscopic geometry~\cite{Huynh_2018_CVPR} or comprehensive skin reflectance~\cite{Yamaguchi2018HFR,sfsnetSengupta18}.
% \haiwei{Not sure if it is good practice to bias towards works from our own lab in citations, although this is understandable.}

% \haiwei{Also, should we consider citing less 3D works and citing more works in 2D domain? } \\

% \haiwei{I think the more relevant works here are: facial image classification (VGGface), landmark detections, and some traditional image synthesis approaches (e.g. inpainting, image morphing) before we move to the more recent generative networks. Maybe we could have a separate subsection for them after related works to face capturing. }

The reconstructed 3D mesh can be used to re-render animated faces using single-view face tracking for video dubbing~\cite{Dale:2011:VFR,GVSSVPT15}, realistic facial reenactment~\cite{thies2015realtime,thies2016face2face}, facial replacement~\cite{NMTHM:2017:face}, or lip-syncing~\cite{suwajanakorn2017synthesizing}. Previous work also explored the techniques for animating a photorealistic avatar by building person-specific skin deformation and texture models from RGB-D~\cite{casas2016rapid} or RGB scans~\cite{cao2016real}. 
In 3D-based methods, 3D face modeling and retargeting from a single-view input~\cite{Bouaziz:2013:OMR,cao2014facewarehouse,garrido2016reconstruction,hsieh2015unconstrained,hu2017avatar,li2013realtime,saito2016realtime,weise2011realtime,li2010example} can be performed to properly decompose person-specific facial shapes from expressions and pose. Alternatively, 2D facial landmarks and image-space detail transfer can be used to animate still portraits~\cite{elor2017bringingPortraits}.

%\paragraph{Deep Generative Methods for Facial Manipulation.}
%\vspace*{-8pt}
\paragraph{Deep Learning-based Methods.}
For still portrait synthesis, GANs~\cite{goodfellow2014generative} have been extensively studied for synthesizing a high-resolution human face, including person-specific details such as pores and facial hair~\cite{karras2017progressive,Karras:2018:stylegan}. 
% For still portrait synthesis, Karras et al.~\cite{karras2017progressive} showed a multi-scale approach to synthesize high-resolution human face by progressively improve the resolution of the GAN~\cite{goodfellow2014generative} synthesis. Later, Karras et al.~\cite{Karras:2018:stylegan} proposed an alternative architecture using a style-based generator to allow high-resolution facial attributes editing (e.g. facial identity and hair). 
A conditional GAN~\cite{isola2016image} has been introduced for pixel-to-pixel translation applications to manipulate a human face image from edge drawings~\cite{wang2018pix2pixHD} or image sequences~\cite{wang2018vid2vid}.
%Isola et al.~\cite{isola2016image} introduced a conditional GAN for pixel-to-pixel translation applications to synthesize a face image from an image of edge drawings~\cite{wang2018pix2pixHD} or a sequence~\cite{wang2018vid2vid}. 
%CycleGAN~\cite{zhu2017unpaired} was introduced to allow training of image translation between two domains using a cycle consistency loss. 
For facial expression editing, Choi et al.~\cite{choi2017stargan} extended previous work to multi-domain pixel translations, enabling facial expression editing using discrete expression labels. Pumarola et al.~\cite{Pumarola_ijcv2019} showed that a conditional GAN with a cycle consistency loss \cite{zhu2017unpaired} can be used for unsupervised learning of continuous facial animation editing from a single image.  However, the controls provided are too coarse to capture the fine-scale nature of human facial expressions. 
For face swapping applications, "DeepFake" frameworks (\textit{e.g.}~\cite{deepfakes2019}) employ an encoder and decoder architecture to achieve a video-based face swapping of a pair of subjects. However, the framework cannot handle arbitrary pairs of subjects without additional subject-specific training. 
For many-to-many subject face swapping, Bao et al.~\cite{Bao_2018_CVPR} proposed a GAN-based training framework to decompose facial identity from other attributes such as expression, pose and illumination, allowing an end-to-end face swapping for unseen subjects. 

An alternative approach for decomposing facial identity from other attributes is to explicitly model it as facial geometry such as 2D landmarks or 3D face models. For 2D geometry-guided methods, Natsume et al.~\cite{natsume18fsnet} proposed a framework to achieve single-image face swapping between unseen identities conditioned on 2D landmarks. Nirkin et al.~\cite{nirkin2019fsgan} proposed a recursive approach for improved identity preservation for a subject-agnostic face image synthesis. Siarohin et al. \cite{NIPS2019FirstOrder} introduced a first order motion model which can animate an image of a variety of categories via keypoints and local affine transformations including a human face portrait. However, it is challenging to properly separate person-specific identity, facial expressions and pose from coarse 2D landmarks (typically 68 fiducial points), and thus the previous work can still suffer from noticeable artifacts and identity mismatches. Zakharov et al.~\cite{zakharov2019few} relaxed the requirement for one-shot learning, thereby showing that few-shot learning could be employed to improve the identity preservation for portrait reenactment. However, it cannot adapt the landmarks when the source and target subjects are different and does not address cross-subject face reenactment. Wang et al. \cite{wang2019fewshotvid2vid} proposed a few-shot framework for general video-to-video translation applications and applied it to animating portraits. Unlike any of the above methods, our method only requires a single image of the target, it does not require subject-specific training, and it can handle cross-subject reenactment of unseen subjects (\textit{i.e.}, it is subject-agnostic).

For 3D geometry-based methods, Kim et al. proposed a hybrid approach combining 3D morphable models and an image translation network to translate 3D rendering of the target face to a synthetic video for video portrait reenactment~\cite{kim2018DeepVideo} or visual dubbing~\cite{Kim19NeuralDubbing} between pairs of subjects. Nagano et al.~\cite{Nagano:2018:PRA} proposed a generalized solution that can synthesize arbitrary expressions of an unseen identity from a single picture, but only operates in the face region. Previous work~\cite{geng2018warp,Geng_CVPR19} also addressed identity-agnostic face image synthesis using 3D face fitting and deep neural nets, but also addressed full portrait manipulation including the background using background warping~\cite{geng2018warp} or blending~\cite{Geng_CVPR19}. 

% An alternative approach that does not require 3D face fitting is 2D landmark-guided synthesis. Natsume et al.~\cite{natsume18fsnet} proposed a framework to achieve single-image face swapping between unseen identities using 2D landmarks as conditions. Later Nirkin et al.~\cite{nirkin2019fsgan} proposed a recursive approach for improved identity preservation for a subject agnostic face image synthesis. However, it is challenging to properly separate person-specific identity, facial expressions and pose from a bunch of 2D landmarks and thus the synthesized face can still suffer from noticeable artifacts. If we relax a requirement on the zero-shot learning, Zakharov et al.~\cite{zakharov2019few} and Wang et al.~\cite{wang2019fewshotvid2vid} showed that few-shot learning could be employed to improve facial identity preservation for face image synthesis. Unlike any of the above methods, our method can preserve person-specific facial shapes and can synthesize realistic face images for unseen subjects (subject agnostic). 
%We demonstrate the disentanglement performance of our approach on a face landmark dataset and evaluate our 

% -Compared to 2D landmark-based method (Few-shot\cite{zakharov2019few}, FSGAN\cite{nirkin2019fsgan}): decouple expression and identity, avoid personality mismatch.
%FSNet\cite{natsume18fsnet}

%-Compared to DeepFake and others (Live Face De-Identification in Video): one-shot, better generability, face shape preserved.

\section{Method}

\begin{figure}[t]
	\footnotesize
	\centering
	\includegraphics[width=0.55\linewidth]{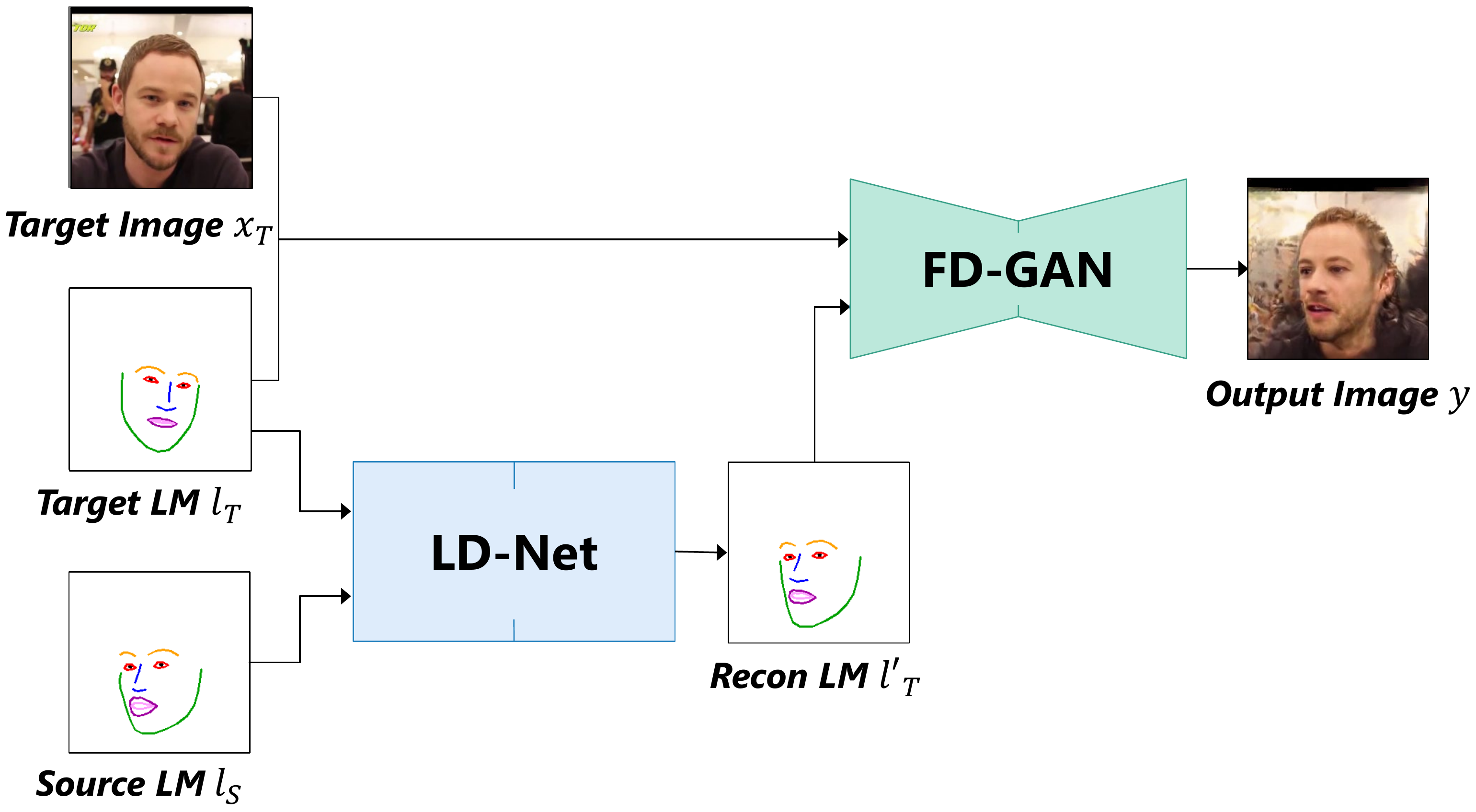}
	%\vspace*{-5pt}
	\caption{Overview of our method, which consists of two sub-networks: LD-Net and FD-GAN. Given the landmarks of a pair of face images $l_T$ and $l_S$, LD-Net first generates a landmark $l'_T$ combining the target's identity with the source's pose/expression. Then, taking $l_T$, $l'_T$ and the target image as input, FD-GAN generates a new face image $y$.}
	\label{fig:pipeline}
	%\vspace*{-10pt}
	\vspace*{-10pt}
\end{figure}

Our goal is to transfer the head pose and facial expression from a source video of one person to a target image of another subject while preserving the target's identity. Based on the observation that 2D facial landmarks contain information about the pose and expression as well as person-specific identity features (\textit{e.g.} the size, shape, proportion, and layout of the facial features), we propose to disentangle the identity and pose/expression from the landmarks and use them for landmark synthesis. As shown in Fig.~\ref{fig:pipeline}, our method consists of two sub-networks. The \textit{Landmark Disentanglement Network} (LD-Net) first synthesizes new landmarks with the target's identity and the source's pose/expression. Then the \textit{Feature Dictionary-based Generative Adversarial Network} (FD-GAN) takes both target and synthetic landmarks as input and translates them into a new face image.
%\vspace*{-10pt}
\subsection{Landmark Disentanglement Network (LD-Net)}
\label{D-Net}

Disentangling landmarks into identity and pose/expression is difficult due to the lack of accurate numerical labeling for pose/expression. Inspired by~\cite{anonymous2020disentangling}, which can disentangle two complementary factors of variations with only one of them labeled, we propose a landmark disentanglement network (LD-Net) to disentangle identity and pose/expression using data with only the subject's identity labeled. More importantly, our network generalizes well to novel identities (\textit{i.e.}, those unseen during training), unlike previous works (\textit{e.g.}~\cite{anonymous2020disentangling}).

\begin{figure}[t]
	\footnotesize
	\centering
	\includegraphics[width=1.0\linewidth]{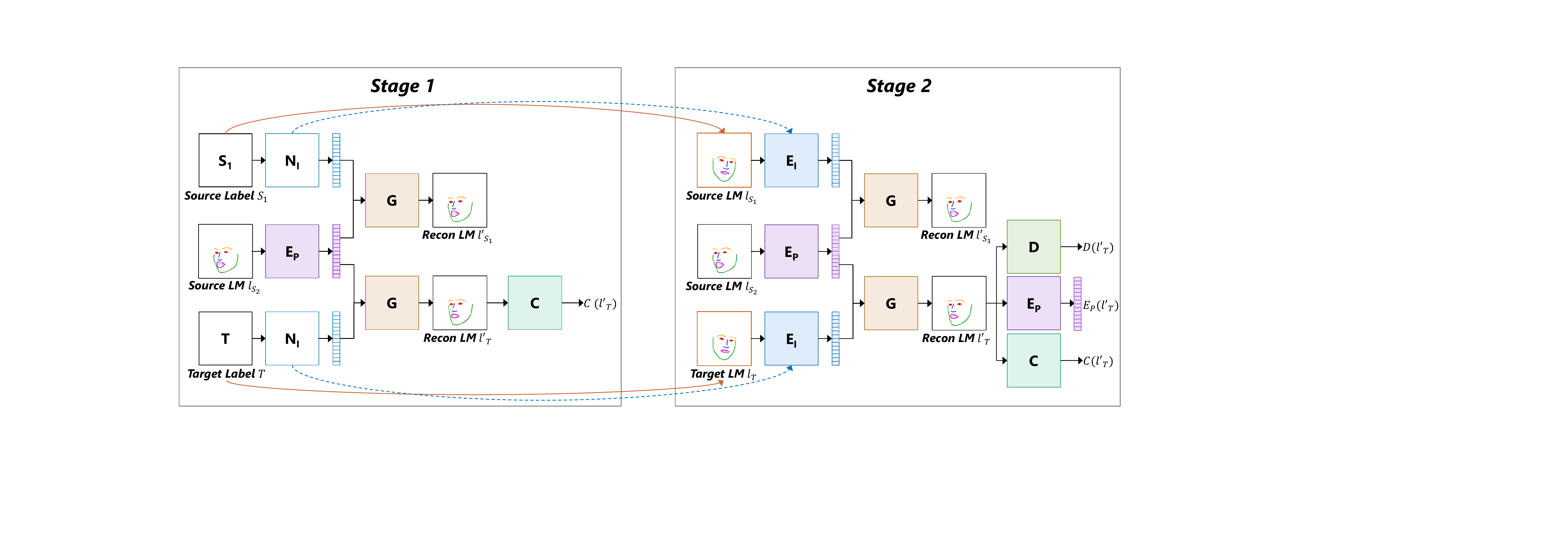}
	%\vspace*{-15pt}
	\caption{The training procedure of LD-Net, consisting of two stages.}
	\label{fig:network1}
	\vspace*{-10pt}
%\vspace*{-15pt}
\end{figure}

Given 2D facial landmarks from a pair of face images, LD-Net first disentangles the landmarks into a pose/expression latent code and an identity latent code, then combines the target's identity code with the source's pose/expression code to synthesize new landmarks. As shown in Fig.~\ref{fig:network1}, the training procedure of LD-Net is divided into two stages. \textit{Stage 1} aims to train a stable pose/expression encoder and \textit{Stage 2} generalizes to predict an identity code from landmarks instead of using identity labels so as to handle unseen identities.
%\todo{@Sitao: why must include Stage 1? Can we provide a reason? What happens if we skip this? } 

\paragraph{Stage 1.} In \textit{Stage 1}, similar to \cite{anonymous2020disentangling}, the network consists of four modules: (1) a pose/expression encoder $E_P$ that computes a code from the input landmarks $l$ that encodes only pose/expression without information about identity; (2) a one-layer network $N_I$ that maps the input one-hot identity label $k$ to an identity code; (3) a generator $G$ that combines a pose/expression code and an identity code to reconstruct landmarks $l'=G(E_P(l),N_I(k))$; and (4) a classifier $C$ that tries to classify the generated landmarks based on their identity.

As shown in Fig.~\ref{fig:network1}, \textit{Stage 1} is trained with two branches for each iteration, which share the same generator $G$ but are associated with their own input and output. In the first branch, the input $k=S_1$ (identity label) and $l=l_{S_2}$ (landmark locations) are from the same subject so the reconstructed landmarks $l'=l'_{S_1}$ should be as same as the input $l$, which can be used to define a reconstruction loss. In the second branch, the input $k=T$ and $l=l_{S_2}$ are from different subjects and the reconstructed landmarks $l'=l'_T$ should contain no information about the identity of $S_1$.

To achieve this, the classifier $C$ tries to classify $l'=l'_T$ as being a landmark of $S_1$ while the pose/expression encoder $E_P$, generator $G$ and identity encoder $N_I$ tries to prevent the classifier from doing so. $C$ is trained with the classification loss of the form:
\begin{equation}
\label{eqn:loss_s11_c}
\mathcal{L}_C={\mathbb{E}}[-\log P(S_1|l')].
\end{equation}

Meanwhile, $E_P$, $G$ and $N_I$ jointly optimize the reconstruction loss minus the identity classification loss as in Eq.~\ref{eqn:loss_s11_g}. The reconstruction loss is defined as per-point squared Euclidean distance using landmark coordinates.
\begin{align}
\label{eqn:loss_s11_g}
\mathcal{L}_G=\lambda_{\textrm{rec}}{\mathbb{E}}[||l-l'||_2^2]+\lambda_C\mathbb{E}[\log P(S_1|l')],
\end{align}
where $\lambda_{\textrm{rec}}=1000$ and $\lambda_C=0.1$. 

%  Thus, only a classification loss is used to prevent the pose/expression encoder $E_P$ from encoding any identity information from $l_{S_2}$:
% \begin{equation}
% \label{eqn:loss_s11_c}
% \mathcal{L}_C={\mathbb{E}}[-\log P(k|l')].
% \end{equation}

All expectations are taken over $l\sim p(l),k \sim p(k)$ where $p(l)$ and $p(k)$ are training distributions of landmarks and identities where $l$ and $k$ may be from different subjects. Different from the network architecture in \cite{anonymous2020disentangling}, all convolutional networks are replaced with Multi-layer Perceptrons (MLP) in LD-Net, since instead of images we operate on landmark coordinates.

\paragraph{Stage 2.} To generalize to novel subjects, we introduce an identity encoder to \textit{Stage 2}. A notable deficiency of \cite{anonymous2020disentangling} is that it does not include any encoder for the labeled data (i.e., identity in our task) and thus it is limited to generating new samples only for the labeled classes in the training data. In~\textit{Stage 2}, we replace the one-layer network $N_I$ with a full-fledged identity encoder $E_I$ that accepts landmarks as input and encodes them into an identity code.

The full training network of \textit{Stage 2} is shown in Fig.~\ref{fig:network1}, also involving two branches similar to \textit{Stage 1}, that is, $l_{S_1}$ and $l_{S_2}$ are from the same subject while $l_{S_2}$ and $l_{T}$ belong to different subjects.

For the second input $l_{S_2}$ and $l_T$ in the second branch, due to unavailability of ground truth, we train a discriminator $D$ and a classifier $C$ to constrain the reconstructed landmarks $l'_T$. We use least square loss for the discriminator $D$ following \cite{mao2017least} to minimize:
\begin{equation}
\label{eqn:loss_s12_d}
\mathcal{L}_D={\mathbb{E}}[(D(l_{S_2})-1)^2+(D(l'_T)+1)^2].
\end{equation}
The classifier $C$ is trained with an adversarial loss on both input landmarks $l_T$ and generated landmarks:
\begin{align}
\mathcal{L}_C={\mathbb{E}}[-\log P(k|C(l_T))]+{\mathbb{E}}[-\log(1-P(k|C(l'_T)],
\end{align}
with expectations taken over $l_{S_2},{l_T}\sim p(l)$, $k\sim p(k)$. $k$ is the identity label of $l_T$. In addition, a content consistency loss term is defined between the generated pose/expression code and its ground truth $E_P(l_{S_2})$:
\begin{align}
\mathcal{L}_\textrm{cont}&={\mathbb{E}}[||E_P(l_{S_2})-E_P(l'_T)||_2^2].
\end{align}

% The input landmarks $l_{S_1}$ and $l_{S_2}$ in the first branch are from the same subject with different pose/expressions while $l_{S_2}$ and $l_{T}$ in the second branch belong to different subjects. The pre-trained pose/expression encoder $E_P$ is fixed in this stage so that it will not encode any identity information of $l_{S_2}$. 

For the first input $l_{S_1}$ and $l_{S_2}$, the reconstructed landmarks $l'_{S_2}$ should have the same pose/expression and identity as $l_{S_2}$. Thus, a reconstruction loss is defined to discourage $E_I$ to encode pose/expression: 
\begin{equation}
\label{eqn:loss_rec2}
\mathcal{L}_{\textrm{rec}}={\mathbb{E}}[||l_{S_2}-l'_{S_2}||_2^2],
\end{equation}
with the expectation over $l_{S_1}$, $l_{S_2}\sim p(l|k)$, and $k\sim p(k)$. $l_1$ and $l_2$ are landmarks from the same subject $k$.

Thus, the identity encoder $E_I$ and generator $G$ are jointly optimized to minimize a weighted sum of the above losses:
\begin{align}
\label{eqn:loss_s12_g}
\mathcal{L}_G&=\lambda_{\textrm{rec}}\mathcal{L}_{\textrm{rec}}\\
&+\lambda_{\textrm{cont}}\mathbb{E}[||E(l_{S_2})-E(l'_{S_1})||_2^2]\nonumber\\
&+\lambda_D\mathbb{E}[D(l'_T)^2]\nonumber\\
&+\lambda_C\mathbb{E}[-\log P(k|C(l'_T))]\nonumber,
\end{align}
where $\lambda_{\textrm{rec}}=1000$, $\lambda_{\textrm{cont}}=0.01$, $\lambda_D=0.1$, and $\lambda_C=0.1$.

\subsection{Feature Dictionary-based Generative Adversarial Network (FD-GAN)}
%\vspace{-6pt}
With the predicted landmarks rasterized into a landmark image, our next goal is to translate it to a photorealistic face image. We can think of the translation procedure as follows: a local patch around each location in the landmark image indicates ``which facial part should be here'', and for each location, we want to translate this into ``how it should appear''. Thus, we propose a novel feature dictionary-based generative adversarial network (FD-GAN) to achieve these intuitive objectives.

\begin{figure*}[t]
	\footnotesize
	\centering
	\includegraphics[width=0.90\linewidth]{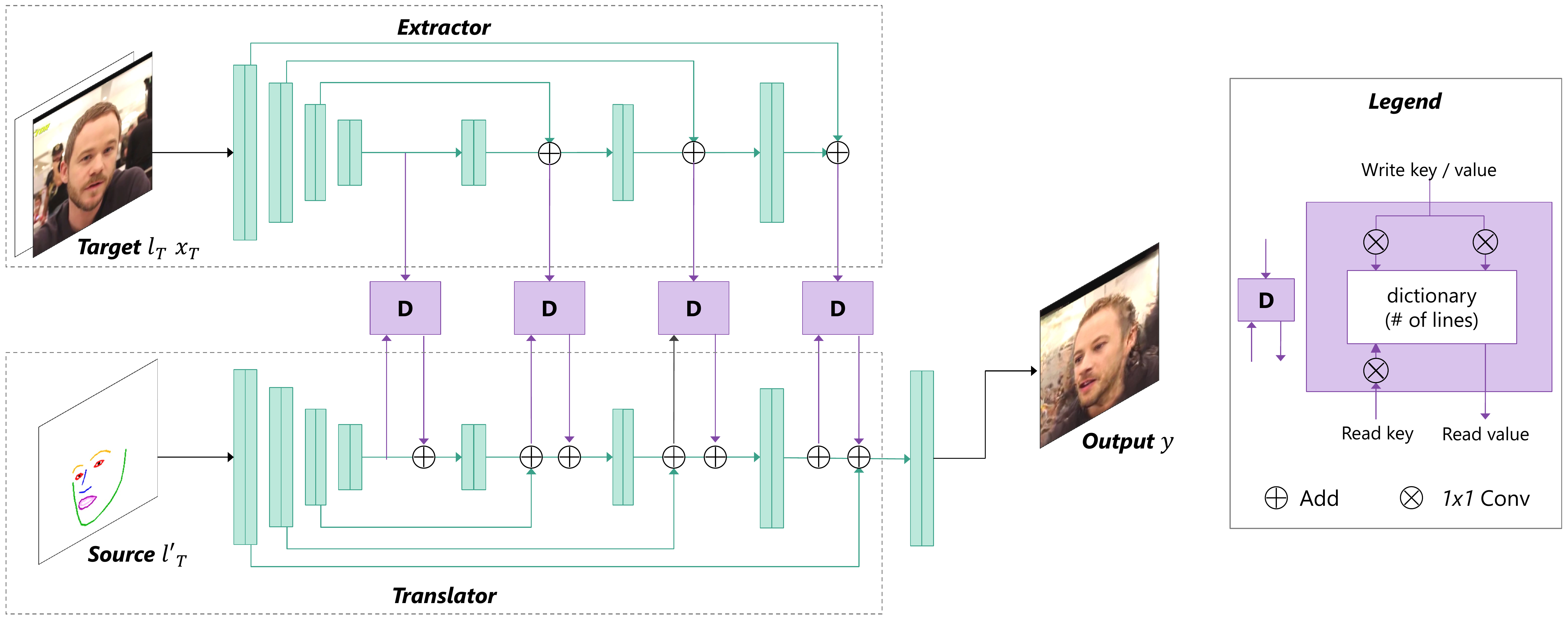}
	%\vspace*{-5pt}
	\caption{The network architecture of the second sub-network FD-GAN.}
	\label{fig:network2}
	%\vspace*{-15pt}
	\vspace*{-10pt}
\end{figure*}
	
The architecture of FD-GAN is illustrated in Fig.~\ref{fig:network2}, which consists of an extractor and a translator. Given a target image $x_T$ and its corresponding landmark image $l_T$, we train an extractor that constructs a ``feature dictionary'' in the module D, which is essentially a mapping from an annotation in the landmark image to its appearance in the target image. Concurrently, given another landmark image $l'_{T}$ and the feature dictionary, we train a translator that retrieves relevant facial features from the dictionary based on the landmarks and composes a face image.

The dictionary mapping is realized with a mechanism similar to the memory bank in the Neural Turing Machine (NTM)~\cite{graves2014neural}: the feature dictionary is a memory matrix, with each memory row conceptually corresponding to some facial component and the value stored in that row corresponding to how that component should appear on a specific subject's face. The construction of such a feature dictionary corresponds to the write operation in NTM and the lookup step during translation corresponds to the read operation in NTM. 
%Our mechanism is simpler in that it is write-once, read-once. 
Precisely, a feature dictionary consists of $n$ rows, and each row $i$ is associated with a write tag $\mathbf{t}^{(i)}$  and read tag $\mathbf{u}^{(i)}$, both of which are vectors of length $m_T$ and are learnable parameters of the network, and a stored value $\mathbf{v}^{(i)}$, which is a vector of length $m_V$ computed by the network.

During the writing phase in the extractor, the stored values are computed as follows: in the form of two convolutional feature maps, the extractor generates for each spatial location $j$ a write key $\mathbf{p}^{(j)}$ and a write value
$\mathbf{x}^{(j)}$. Then, the value of $\mathbf{v}^{(i)}$ is computed as:
\begin{equation}
\mathbf{v}^{(i)}=\frac{\sum_{n\in N}\mathbf{x}^{(n)}\cdot e^{({\mathbf{t}^{(i)} \cdot \mathbf{p}^{(n)}})}}{\sum_{n\in N}e^{({\mathbf{t}^{(i)} \cdot \mathbf{p}^{(n)}})}},
\end{equation}
where $N$ is the set of spatial locations. That is, the value of row $i$ in the feature dictionary is a weighted sum of the write values at each location, with weight being the softmax of $\mathbf{t}^{(i)} \cdot \mathbf{p}^{(n)}$ for location $n$.

Similarly, during the reading phase, the translator, as a convolutional feature map, generates for each spatial location
$j$ a read key $\mathbf{q}^{(j)}$. The return value $\mathbf{y}^{(j)}$ for each location of lookup operation is computed as:
\begin{equation}
\mathbf{y}^{(j)}=\frac{\sum_{k=1}^{n}\mathbf{v}^{(k)}\cdot e^{\mathbf{u}^{(k)} \cdot \mathbf{q}^{(j)}}}{\sum_{k=1}^{n}e^{\mathbf{u}^{(k)} \cdot \mathbf{q}^{(j)}}},
\end{equation}
which means the value read by each location $j$ is a weighted sum of all the rows in the feature dictionary, with weight being the softmax of $\mathbf{u}^{(k)} \cdot \mathbf{q}^{(j)}$ for each row $k$. The translator then continues network operations on this returned convolutional feature map.

The extractor and translator are both fully convolutional, with U-Net skip connections as shown in Fig.~\ref{fig:network2}. 
%The feature dictionaries are multi-scale: the extractor writes a separate feature dictionary at each different scale, and each is read by the translator at the same scale. The number of lines, tag length and value length can be independently chosen for each scale.
To train such a joint extractor-translator, we employ a combination of reconstruction loss, GAN loss, and an adversarial classifier loss. The discriminator and classifier are both patch-based with their loss averaged across spatial locations.
In the following equations, $T$ is the extractor-translator, and to avoid excessive notation, we use the same letters $D$ and $C$ for the discriminator and classifier as in Sec.~\ref{D-Net}. 
%They are also optimized with very similar loss functions, but otherwise has no connection to their counterpart in Section~\ref{D-Net}.
For simplicity, we omit the range over which the expectations are taken: $x_T$ and $x'_T$ are two frames from the same video clip, $l_T$ and $l'_T$ are their respective landmark images, and
$k$ is the identity label of $x_T$ and $x'_T$.

The discriminator $D$ minimizes:
\begin{equation}
\mathcal{L}_D={\mathbb{E}}[(D(x_T)-1)^2+(D(T(x_T, l_T, l'_T))+1)^2].
\end{equation}

The classifier $C$ minimizes:
\begin{align}
\mathcal{L}_C&={\mathbb{E}}[-\log P(k|C(x_T))]\\
&+{\mathbb{E}}[-\log(1-P(k|C(T(x_T, l_T, l'_T))))]\nonumber.
\end{align}

The loss of extractor-translator in FD-GAN is a weighted sum of adversarial discriminator loss, adversarial classifier loss and reconstruction:
\begin{align}
\mathcal{L}_T&=\lambda_{\textrm{rec}}{\mathbb{E}}[||T(x_T, l_T, l'_T)-x'_T||_2^2]\\
&+\lambda_D{\mathbb{E}}[D(T(x_T, l_T, l'_T))^2]\nonumber\\
&+\lambda_C{\mathbb{E}}[-\log P(k|C(T(x_T, l_T, l'_T)))]\nonumber,
\end{align}
where $\lambda_{\textrm{rec}}=50$, $\lambda_D=1$, and $\lambda_C=1$.

\section{Experiments}
% \begin{enumerate}
%     \item \todo{In figures, images and text are misaligned, hard to follow.}
%     \item \todo{The contributions need to be justified by all the experiments.}
%     \item \todo{The proposed disentanglement method further degrades the quality of the results, as can be seen from Table 3.}
%     \item \todo{More qualitative comparison. (The proposed method compares with [58] qualitatively, since the numerical comparison is clearly in favor of [58].)}
%     \item \todo{More quantitative results on challenging cases.}
%     \item \todo{Time table: to show a timing comparison of the approach 
% 34
%  against other approaches.}
%     \item \todo{LandmarkTest: where does this data come from? Most importantly, how exactly are the pairs with same expression but different identity obtained?}
% \end{enumerate}{}

We first conduct an evaluation and ablation study in Sec.~\ref{sec:ablation_study} on the performance of LD-Net and FD-GAN independently, followed by comparisons of our full method with the state-of-the-art methods on cross-subject face reenactment in Sec.~\ref{sec:comparison}. For more results tested on unconstrained portrait images, please refer to the supplemental material.

\paragraph{Implementation details.} %In LD-Net, all networks are MLPs. 
For FD-GAN, the extractor and translator are based on U-Nets, with both networks joined together by dictionary writer/reader modules inserted into the up-convolution modules. The discriminator and classifier for FD-GAN are patch-based and have the same structure as the down-convolution part of the U-Nets. Please refer to the supplemental material for more details concerning the network structures and training strategies. 

\paragraph{Performance.} Our method takes approximately 0.08s for FD-GAN to generate one image and 0.02s for LD-Net to perform landmark disentanglement on a single NVIDIA TITAN X GPU.

\paragraph{Training datasets.} The training dataset is built from VoxCeleb video training data~\cite{chung2018voxceleb2} which is processed by dlib~\cite{dlib09} to crop a 256$\times$256 face image at 25fps and to extract its landmarks. In total, it contains 52,112 videos for 1,000 randomly selected subjects. 
%LD-Net is in principle independent of image size, so we normalize landmark coordinates to the range [-1, 1]. 

\paragraph{Testing datasets.} We use three datasets to evaluate our method:
%\vspace{-.1in}
\begin{itemize}
\setlength{\itemindent}{.15in}
    \item[$\bullet$] LMTest: a landmark dataset which has 200,000 landmarks (100 subjects $\times$ 2000 frames of varying poses and expressions) with ground truth labels for both identity and poses/expressions. Using a video of one person performing and the first 100 neutral expression photos from the Compound Facial Expressions Database \cite{CFE2014}, we used single-view 3D face fitting \cite{thies2016face2face} to retarget facial expressions and poses from the video subject to each subject's 3D face model and project 3D vertex positions to obtain ground truth 2D landmarks. This dataset is used to evaluate the effect of LD-Net.
    \item[$\bullet$] SelfTest: a video dataset of 8,000 frames for 80 subjects from Voxceleb testing data (100 frames per video at 25fps). It is only used for the ablation study when testing self-reenactment, where the ground truth is known. 
    \item[$\bullet$] CrossTest: a video dataset of 8,000 frames for 80 pairwise subjects (100 frames per video at 25fps) randomly sampled from the Voxceleb testing data, used to compare our method with the baselines in one-shot cross-subject face reenactment.
\end{itemize}

\noindent\textit{Metrics.} We use the following metrics for quantitative evaluation of generated images.
%\vspace{-.1in}
\begin{itemize}
\setlength{\itemindent}{.15in}
    \item[$\bullet$] Identity Similarity (ISIM): computes cosine similarity between embedding vectors of the face recognition network VGGFace2~\cite{cao2018vggface2} for identity matching.
    \item[$\bullet$] Pose Similarity (PSIM): computes cosine similarity between head rotation in radians around the X, Y, and Z axes estimated by OpenFace~\cite{baltrusaitis2018openface}.
    % \item[$\bullet$] Expression distance (ED): computes L2 distance between expression vectors retrieved by a 3D face model~\cite{chaudhuri2019joint} for measure expression consistency.
    \item[$\bullet$] Expression Distance (ED): computes L2 distance of intensities of corresponding facial action units detected by OpenFace~\cite{baltrusaitis2018openface} between the generated images and the driving images. 
    %(\todo{decide to use ED or ES based on which one can be implemented fast.})
    \item[$\bullet$] Fr{\'e}chet-Inception Distance (FID)~\cite{heusel2017gans}: measures the distance between the distributions of real data and generated data to quantify the result fidelity.
    \item[$\bullet$] Structured Similarity (SSIM): measures low-level similarity to ground truth images in the self-reenactment setting.
\end{itemize}

\subsection{Evaluation}
\label{sec:ablation_study}
\paragraph{Evaluation of LD-Net.} To validate the accuracy in disentangling identity and pose/expression, we test the LD-Net in isolation using the LMTest dataset. From the 200,000 landmarks, we sample pairs of landmarks in 3 different patterns: the same identity but different pose/expression, the same pose/expression but different identity, and both differing identity and pose/expression. In each case, we randomly sample one million pairs of landmarks and compute their distances in the latent space of identity encoder $\IEncoder$ and pose/expression encoder $\PEncoder$ respectively. We first use PCA to reduce the dimensionality of $E_I$'s and $E_P$'s latent codes to 8 before computing the mean Euclidean distance for each pair. For $E_I$'s latent space, pairs of landmarks from the same subject should give a smaller mean distance than pairs from different identities, and similarly for landmarks with the same pose/expression in the latent space of $\PEncoder$. Table~\ref{tab:mean} gives the mean distances for each case, which shows that the identity code and pose/expression code do control the respective aspect of the generated landmarks, no matter what kind of input is provided.
\begin{table}{}
%\vspace*{-5pt}
\centering
    \begin{tabular}{l|c|c}
        \hline
        Sample by & In $E_I$'s space $\downarrow$  & In $E_P$'s space $\downarrow$\\
        \hline
        Same identity & \textbf{2.0431} & 3.5171\\
        Same pose/exp & 3.1793 & \textbf{1.2730}\\
        Both different & 3.8274 & 3.6744\\
        \hline
        \end{tabular}
    \caption{Mean Euclidean distance in the latent spaces of the identity encoder $E_I$ and the pose/expression encoder $E_P$ in different cases.}
    \label{tab:mean}
    \vspace*{-5pt}
\end{table}
%\vspace{-.3in}

\begin{figure}[t]
	\footnotesize
	\centering
	\includegraphics[width=0.99\linewidth]{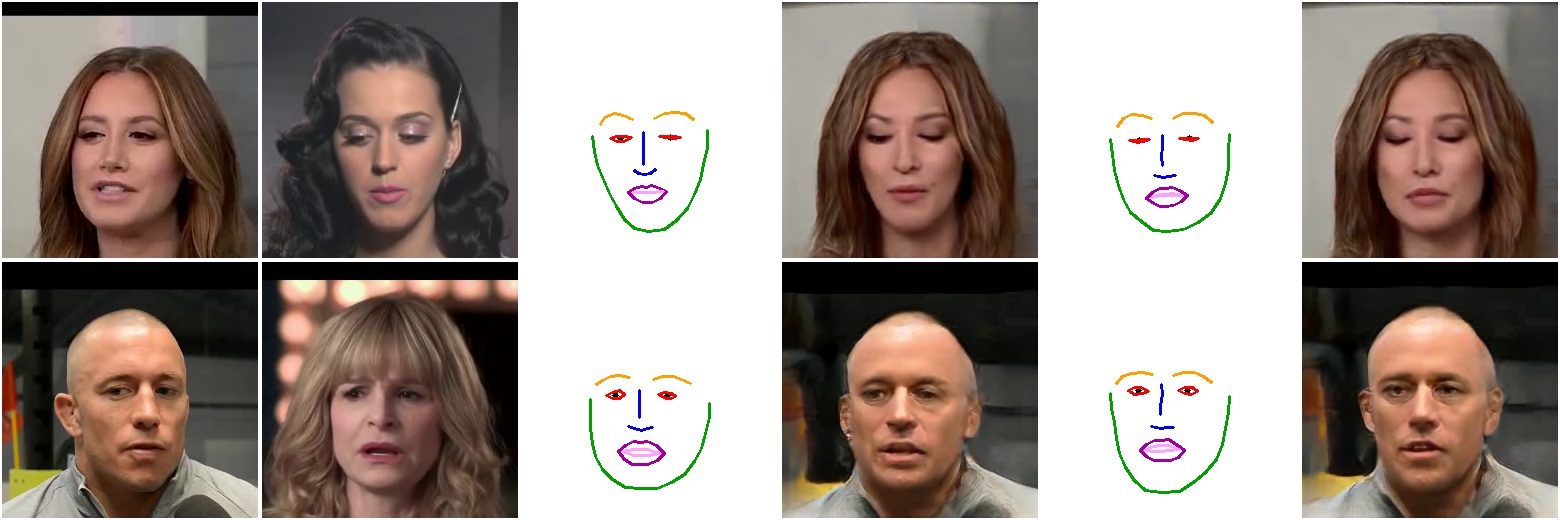}
    % \makebox[0.9\linewidth][s]{
    % 	\makebox[0.145\linewidth]{\scriptsize Target image}
    % 	\makebox[0.145\linewidth]{\scriptsize Source image}
    % 	\makebox[0.29\linewidth]{\scriptsize Source landmarks \&  Result}
    % 	\makebox[0.29\linewidth]{\scriptsize LD-Net landmarks \& Result}
    % }
    \setlength{\tabcolsep}{0\linewidth}
     \begin{tabular}{C{0.166\linewidth}C{0.166\linewidth}C{0.333\linewidth}C{0.333\linewidth}}
     Target & Source & Source landmarks & LD-Net landmarks \\
     image & image & \&  Result & \& Result\\
     \end{tabular}
	\caption{Qualitative comparison between the results generated using source landmarks and synthetic landmarks by LD-Net. Synthesized landmarks better preserve the target identity.}
	\vspace*{-10pt}
	\label{fig:ld_net}
\end{figure}

\paragraph{Ablation analysis of LD-Net.} In Table \ref{tab:ld_net}, We show the effect of LD-Net on the generated images in terms of identity, expression and pose preservation in two settings: self-reenactment and cross-subject. We use three metrics: ISIM, PSIM, and ED to measure matching accuracy of identity, pose and expression, respectively. In self-reenactment, we compare results generated using ground truth landmarks and synthetic landmarks by LD-Net. In cross-subject reenactment, we do a similar comparison between the results using the source subject landmarks as-is and LD-Net landmarks. Since PSIM and ED compare the poses/expressions with the source images, the results using the source landmarks (ground truth) always lead to better matching accuracy. However, the landmarks generated by LD-Net are very close to the ground truth landmarks when comparing in self-reenactment. Moreover, the cross-subject setting shows the importance of predicting personalized landmarks with higher identity accuracy in the cross-subject reenactment in Table.~\ref{tab:ld_net} and better visual quality in Fig.~\ref{fig:ld_net}.
\begin{table}
\centering
    \begin{tabular}{l|c|c|c|c|c|c|c}
    \hline
    & \multicolumn{3}{c|}{Self-reenactment} & & \multicolumn{3}{c}{Cross-subject} \\ 
    \hline
    LM from & ISIM $\uparrow$ & PSIM $\uparrow$ & ED $\downarrow$ & LM from & ISIM $\uparrow$ & PSIM  $\uparrow$ & ED $\downarrow$ \\ 
     \hline
    Ground truth & \textbf{0.7986} & \textbf{0.9134} & \textbf{0.1296} & Source & 0.7145 & \textbf{0.8615} & \textbf{0.2080}\\
    LD-Net & 0.7984  & 0.8950 & 0.1655 & LD-Net & \textbf{0.7726} & 0.8398 &0.2430\\
    \hline
    \end{tabular}
    \caption{Quantitative comparison between the results generated using ground truth landmarks vs landmarks predicted by LD-Net in self-reenactment, and using landmarks from the source subject as-is vs landmarks by LD-Net in cross-subject reenactment.}
    %\vspace*{-10pt}
    \vspace*{-5pt}
    \label{tab:ld_net}
 \end{table}
 %\vspace*{-15pt}
\vspace{-.3in}

\paragraph{Ablation analysis of FD-GAN.} We construct three baselines to evaluate the performance of FD-GAN: Pix2PixHD~\cite{wang2018pix2pixHD}, FD-GAN-1, and AdaIN. The first, Pix2PixHD~\cite{wang2018pix2pixHD}, is an advanced image-to-image translation network which can synthesize photo-realistic images from landmark images. In FD-GAN-1, we reduce the number of rows in the feature dictionary to 1 with its value being the mean of the convolution features. The final baseline, AdaIN, also uses a one-row dictionary but uses the written value to generate parameters for adaptive instance normalization (AdaIN)~\cite{huang2017adain}. 

We compare FD-GAN with the three baselines in both the self-reenactment and cross-subject settings. To evaluate FD-GAN alone, we utilize two important image quality metrics, SSIM (unavailable in cross-subject reenactment) and FD, in addition to ISIM. As shown in Table.~\ref{tab:compare_voxceleb}, the quantitative comparison in both settings demonstrates that our FD-GAN best preserves low-level image features, image fidelity, and identity information. Since our generative network learns a local mapping between the target image and landmarks to the final image, it is flexible enough to generalize to unseen subjects. But existing image-to-image approaches such as Pix2PixHD~\cite{wang2018pix2pixHD} lack the domain generalization
capability needed to synthesize unseen subjects without any subject-specific learning.
%In order to evaluate components of FD-GAN alone, we use the ground truth landmarks as input to generate face images in this experiment. 
%Independent of LD-Net, we use the ground truth landmarks as source and the first frame and its corresponding landmark as the target to generate the face images. 
% The quantitative comparison with variants of our method in Table.~\ref{tab:compare_voxceleb} shows that our FD-GAN better preserves image features and identity information. Since our generative network learns a local mapping between the input/landmarks to the final image, it has better flexibility to generalize to unseen subjects without person-specific learning. 
 \begin{table}{}
\centering
    \begin{tabular}{l|c|c|c|c|c|c}
    \hline
        & \multicolumn{3}{c|}{Cross-subject} & \multicolumn{3}{c}{Self-reectment} \\ 
        \hline
        Method &  ISIM $\uparrow$ & SSIM $\uparrow$ & FID $\downarrow$ & ISIM $\uparrow$ & SSIM $\uparrow$ & FID $\downarrow$\\
        \hline
        Pix2PixHD~\cite{wang2018pix2pixHD} & 0.514 & N/A &  99.34 & 0.41 & 0.49 & 98.36  \\
        AdaIN & 0.650 & N/A &  86.69 & 0.51 & 0.56 & 90.70  \\
        FD-GAN-1 & 0.6232 & N/A & 71.11 & 0.49 & 0.56 & 70.34  \\
        FD-GAN (Ours) & \textbf{0.7726} & N/A & \textbf{67.68} & \textbf{0.63} & \textbf{0.63} & \textbf{55.19} \\
        \hline
        \end{tabular}
    \caption{Quantitative comparison of FD-GAN with three baselines in both self-reenactment and cross-subject reenactment.}
    \label{tab:compare_voxceleb}
    %\vspace*{-15pt}
    \vspace*{-5pt}
\end{table}
\vspace{-.3in}

\subsection{Comparison}
\label{sec:comparison}
\paragraph{Comparison with one-shot methods.} 
We first show quantitative comparisons with two state-of-the-art one-shot face reenactment baselines, X2Face~\cite{Wiles_2018_ECCV}, and First-order-model~\cite{NIPS2019FirstOrder}, using their pre-trained models on the Voxceleb training dataset. We evaluate the models in the same setting without any fine-tuning on the CrossTest dataset. Both X2Face and First-order-model are warping-based methods which can well generalize to unseen subjects in the one-shot setting. In X2Face, the generated frame inherits the object proportions of the driving source video, and the quality of their results is very sensitive to the cropping region and face alignment as shown in Fig.~\ref{fig:one-shot} (we also test the algorithm with a different crop size in the supplemental material). From the quantitative comparison in Table~\ref{tab:compare_one} and the qualitative comparison in Fig.~\ref{fig:one-shot}, we can see that the results from the First-order-model demonstrate the best image fidelity, since it uses a warping formulation to generate the deformed faces. However, its warping formulation, which is based on keypoints and local affine transformations, can hardly provide as accurate local control as our synthesized landmarks which better preserve the source pose/expression, especially when handling very different head poses and facial expressions. Therefore, compared to these methods, our model can generalize to unseen subjects with better identity preservation and more consistent quality under a large variety of poses/expressions.

% However, it also suffers from warping distortion when under large head motion and cannot accurately preserve facial expression. Compared to theirs, our model can generalize to unseen subject with better identity and consistent quality under a variety of poses/expressions.
%Meanwhile we compare with the state-of-the-art method of few-shot learning~\cite{zakharov2019few} using their pretrained model on their test set in Voxceleb2. Table~\ref{tab:compare_samsung} shows the compared settings and metrics, from which it shows that our method can achieve comparable and even better quality with zero-shot learning.
% Quantitative comparison with the state-of-the-art methods on FFTest dataset in one-shot cross-subject face reenactment, 2D-based methods including X2face~\cite{Wiles_2018_ECCV}, Pix2pixHD~\cite{wang2018pix2pixHD}, and First-order-model~\cite{Siarohin_2019_NeurIPS}. We want to show that Pix2pixHD and X2Face suffer from severe problem in domain transfer without person-specific fine-tuning. The warping-based methods X2Face and First-order-model cannot handle large motions. Compared to theirs, our model can generalize to unseen subject with better quality in pose/expression, identity and fidelity.
 \begin{table}{}
\centering
    \begin{tabular}{l|c|c|c|c}
        \hline
        Method & ISIM $\uparrow$ & PSIM $\uparrow$ & ED $\downarrow$ & FID $\downarrow$ \\
        \hline
        X2face~\cite{Wiles_2018_ECCV} & 0.6347 & 0.302 & 0.448 & 101.72 \\
        %Pix2pixHD~\cite{wang2018pix2pixHD} & ? & 0.883 & 0.227 & 98.89  \\
        First-order-model~\cite{Siarohin_2019_NeurIPS} & 0.7699 & 0.822 & 0.274 &  \textbf{55.94} \\
        Ours  & \textbf{0.7762} & \textbf{0.840} & \textbf{0.243} & 67.68 \\
        \hline
        \end{tabular}
    \caption{Quantitative comparison of methods for cross-subject reenactment on the CrossTest dataset between our method and~\cite{Wiles_2018_ECCV},~\cite{wang2018pix2pixHD}, and~\cite{Siarohin_2019_NeurIPS}.}
    \label{tab:compare_one}
    \vspace*{-5pt}
    %\vspace*{-15pt}
\end{table}
\vspace{-.3in}

 \begin{figure}[t]
	\footnotesize
	\centering
	\includegraphics[width=0.99\linewidth]{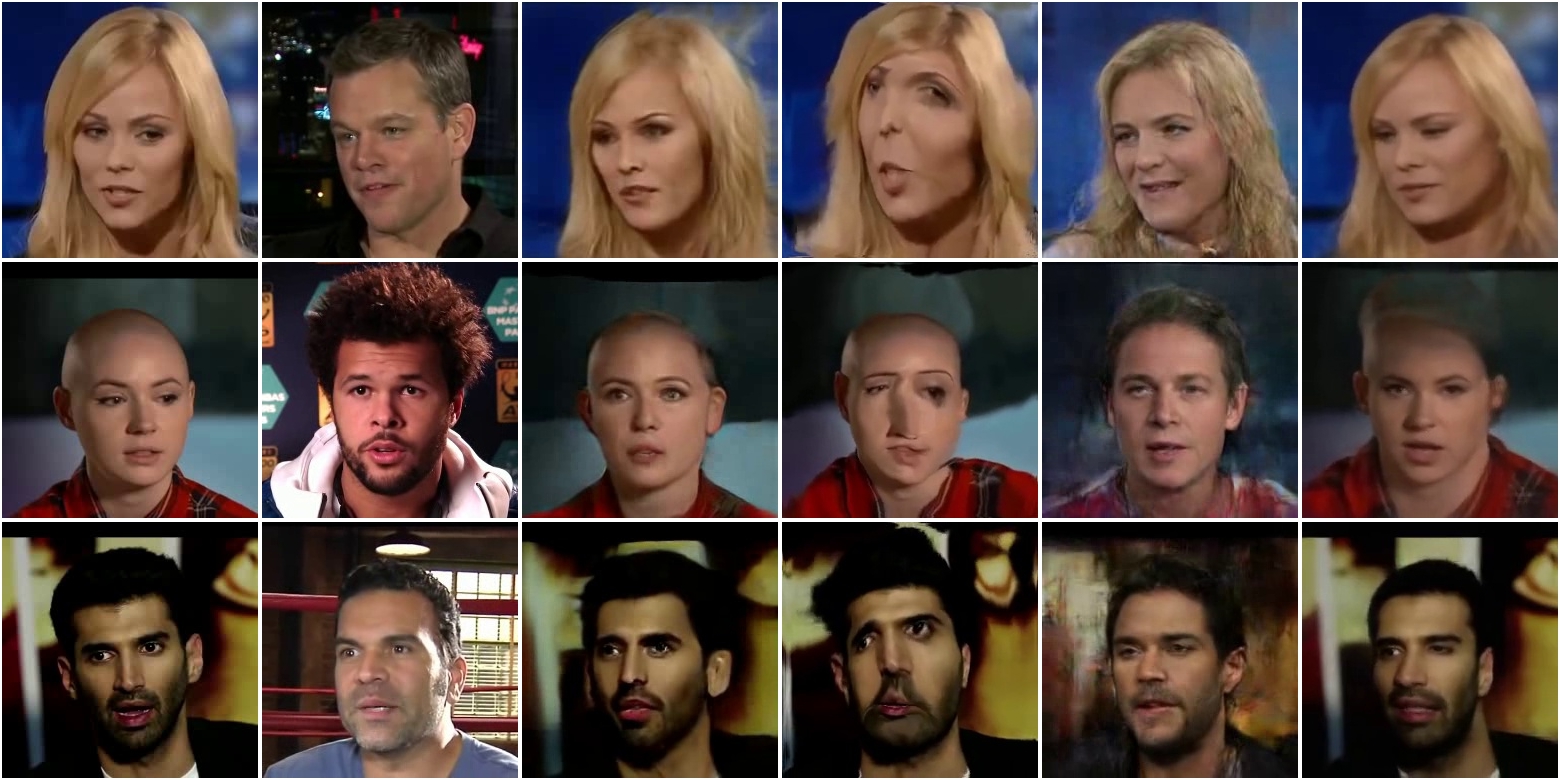}
    % \makebox[0.85\linewidth][s]{
    % 	\makebox[0.135\linewidth]{\scriptsize Target}
    % 	\makebox[0.135\linewidth]{\scriptsize Source}
    % 	\makebox[0.135\linewidth]{\scriptsize Ours}
    % 	\makebox[0.135\linewidth]{\scriptsize X2face}
    % 	\makebox[0.135\linewidth]{\tiny Pix2pixHD}
    % 	\makebox[0.135\linewidth]{\tiny First-order-model}
    % }
    \setlength{\tabcolsep}{0\linewidth}
     \begin{tabular}{C{0.166\linewidth}C{0.166\linewidth}C{0.166\linewidth}C{0.166\linewidth}C{0.166\linewidth}C{0.166\linewidth}}
     Target & Source & Ours & X2face & Pix2PixHD & First-order- \\
     image & image & & \cite{Wiles_2018_ECCV} & \cite{wang2018pix2pixHD} & model \cite{Siarohin_2019_NeurIPS}\\
     \end{tabular}
     
	\caption{Qualitative comparison between our method and the baselines in cross-subject reenactment:  X2face~\cite{Wiles_2018_ECCV}, Pix2PixHD~\cite{wang2018pix2pixHD}, and First-order-model~\cite{Siarohin_2019_NeurIPS}.}
	\label{fig:one-shot}
	\vspace*{-10pt}
\end{figure}

% \paragraph{Qualitative comparison with few-shot methods.} %\todo{Include this comparison or not?} 
% Visually compare with two state-of-the-art few-shot methods~\cite{zakharov2019few} and~\cite{ha2019marionette} on the examples provided by~\cite{ha2019marionette} since their official implementation and pre-trained models are unavailable. We want to show that our method can better preserve identity compared to~\cite{zakharov2019few} and have fewer blurrier artifacts than~\cite{ha2019marionette} even though we use only one target images while they use multiple. Also, although~\cite{ha2019marionette} can also handle one-shot setting but always fails while performing a one-shot reenactment under different identity setting due to large pose difference (please refer to comparison between our method and~\cite{ha2019marionette} in one-shot cross-subject reenactment in supplemental material). 

\begin{figure}[t]
	\footnotesize
	\centering
	\includegraphics[width=0.84\linewidth]{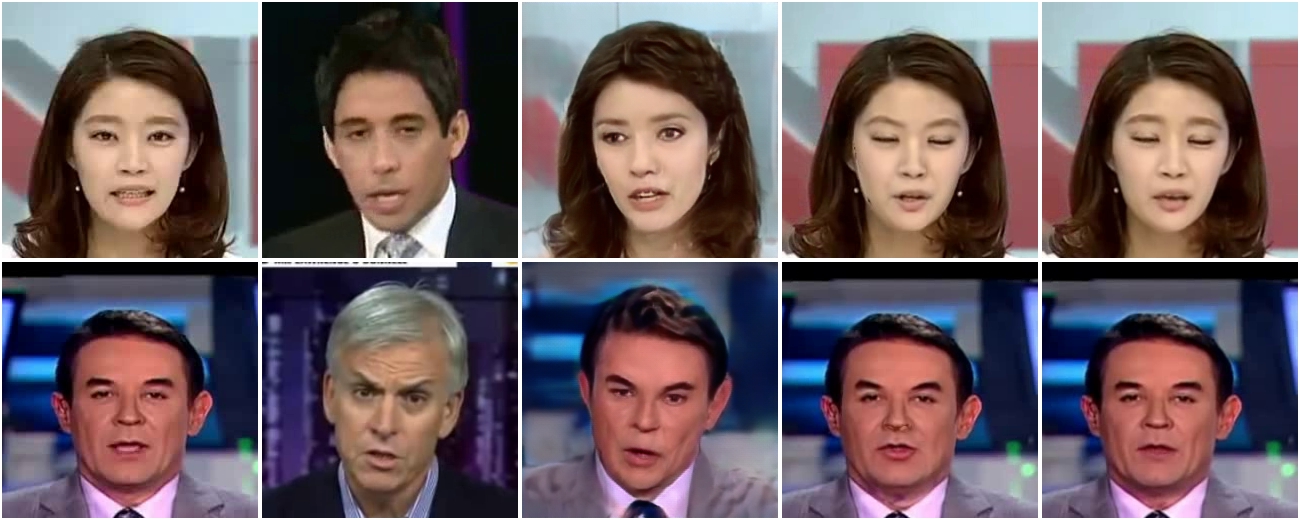}
    % \makebox[0.85\linewidth][s]{
    % 	\makebox[0.16\linewidth]{\scriptsize Target image}
    % 	\makebox[0.16\linewidth]{\scriptsize Source image}
    % 	\makebox[0.16\linewidth]{\scriptsize Ours}
    % 	\makebox[0.16\linewidth]{\tiny Face2Face~\cite{thies2016face2face}}
    % 	\makebox[0.16\linewidth]{\tiny NeuralTexture~\cite{thies2019deferred}}
    % }
    \setlength{\tabcolsep}{0\linewidth}
     \begin{tabular}{C{0.166\linewidth}C{0.166\linewidth}C{0.166\linewidth}C{0.166\linewidth}C{0.166\linewidth}}
     Target & Source & Ours & Face2Face & Neural- \\
     image & image & & \cite{thies2016face2face} & Texture \cite{thies2019deferred}\\
     \end{tabular}
	\caption{Qualitative comparison between our method and two 3D-based methods: Face2Face~\cite{thies2016face2face} and NeuralTexture~\cite{thies2019deferred}.}
	\label{fig:3D}
	\vspace*{-10pt}
\end{figure}

%  \begin{figure}[t]
% 	\footnotesize
% 	\centering
% 	\includegraphics[width=1.0\linewidth]{figure/placeholder.jpg}
% 	\begin{tabular}{ccccc}
% 	Target image & \quad Source image & \cite{zakharov2019few} & \cite{ha2019marionette} & Ours \\
% 	\end{tabular}
% 	\caption{\todo{Qualitative comparison between our method and few-shot methods.}}
% 	\label{fig:few-shot}
% \end{figure}
%\vspace{.3in}
\paragraph{Qualitative comparison with 3D-based methods.} 
%Visually compare with two methods (\ie, Face2Face~\cite{thies2016face2face}, NeuralTexture~\cite{thies2019deferred}) based 3D fitting on FFTest dataset. We want to show that their face textures present higher quality because they directly copy from the target input (@koki: is that right?) but fail to accurately change head poses and expressions. 
Fig.~\ref{fig:3D} shows qualitative comparisons on the FaceForensics++ test dataset~\cite{roessler2019faceforensicspp} with two state-of-the-art 3D-based methods (Face2Face~\cite{thies2016face2face} and NeuralTexture~\cite{thies2019deferred}). 
Compared to their methods which require 3D face fitting to maintain the target identity and cannot change head poses, our method can synthesize personalized faces with arbitrary head poses using only 2D landmarks. %Their face textures present higher quality than ours but fail to change head poses. 

% \subsection{Results}
% (Figure) best results of our method shown on in-the-wild face images.
% \label{sec:results}

%  \begin{figure}[t]
% 	\footnotesize
% 	\centering
% 	\includegraphics[width=1.0\linewidth]{figure/placeholder.jpg}
% 	\begin{tabular}{ccc}
% 	Target image & \quad Source image & & Ours \\
% 	\end{tabular}
% 	\caption{\todo{Our cross-subject face reenactment results on in-the-wild face images under various challenging conditions.}}
% 	\label{fig:wild}
% \end{figure}

%\input{experiment_old.tex}

\section{Discussion and Future Work}
We have demonstrated a technique for portrait reenactment that only requires a single target picture and 2D landmarks of the target and the driver. The resulting portrait is not only photorealistic but also preserves recognizable facial features of the target. Our comparison shows significantly improved results compared to state-of-the-art single-image portrait manipulation methods. Our extensive evaluations confirm that identity disentanglement of 2D landmarks is effective in preserving the identity when synthesizing a reenacted face. We have shown that our method can handle a wide variety of challenging facial expressions and poses of unseen identities without subject-specific training. This is made possible thanks to our generator, which uses a feature dictionary to translate landmark features into a photorealistic portrait.

A limitation of our method is that the resulting portrait has only a resolution of 256$\times$256, and it is still difficult to capture high-resolution person-specific details such as stubble hair. It could also suffer from some artifacts for non-facial parts and the background region, since we rely on the landmarks to transfer facial appearance but the landmarks contain no structural information about the hair or background. We believe such a limitation could be further addressed by incorporating dense pixel-wise conditioning~\cite{Nagano:2018:PRA} and segmentation. While our method can produce reasonably stable portrait reenactment results from a frame of target and 2D landmarks, the temporal consistency could be further improved by taking into account temporal information from the entire video. 

%Our portrait reenactment result could be combined with facial segmentation and blending to achieve a new application such as photorealistic face-swapping~\cite{nirkin2019fsgan}.

\section*{Acknowledgment}
We would like to thank Qingguo Xu for his help on processing the VoxCeleb video dataset and Kyle Olszewski for proofreading this manuscript. This research was conducted at USC and was funded by in part by the ONR YIP grant N00014-17-S-FO14, the CONIX Research Center, a Semiconductor Research Corporation (SRC) program sponsored by DARPA, the Andrew and Erna Viterbi Early Career Chair, the U.S. Army Research Laboratory (ARL) under contract number W911NF- 14-D-0005, Adobe, and Sony. This project was not funded by Pinscreen, nor has it been conducted at Pinscreen or by anyone else affiliated with Pinscreen. Koki Nagano is affiliated with Pinscreen but worked on this project through his affiliation at USC/ICT. The content of the information does not necessarily reflect the position or the policy of the Government, and no official endorsement should be inferred.

\clearpage
\bibliographystyle{splncs04}
\bibliography{egbib}

\clearpage
\section*{Appendix}
In this supplementary material, we first explain details of the implementation, training strategy and performance of our method. We then provide additional results for evaluations and qualitative comparisons between our method and other one-shot face reenactment baselines on different datasets. Finally, we demonstrate the strong capability of our method by testing on in-the-wild portrait images from the Internet. More video results can be found in the supplementary video.

\section{Implementation Details}
All networks are MLPs in LD-Net, having 10
hidden layers with 512 features each. The length of the pose/expression code is 64 and the length of the identity code is 128.

In FD-GAN, the extractor and translator are based on U-Nets, with both networks joined together by dictionary writer/reader modules inserted in the up-convolution modules of the network, as shown in Fig. 3. in the paper.

In the down-convolution module, each level consists of a stride-2 convolution with 4$\times$4 kernel, followed by a flat convolution with a 3$\times$3 kernel. In the up-convolution module of the
extractor, each level consists of a dictionary writer module, followed by a flat convolution with a 3$\times$3 kernel and then a stride-2 convolution with 4$\times$4 kernel. The up-convolution module of the translator is similar, with dictionary writers replaced with readers.

In the writer modules, write keys and write values are each computed from the input with a 1$\times$1 convolution. In the reader modules, read keys are computed from the input with a 1$\times$1 convolution and the values read from the dictionary are added back to the input feature, as in a residual block.

The discriminator and classifier for the generation part are patch-based and have the same structure as the down-convolution module of the U-Nets. For all networks, from the lowest level to the highest, the number of convolutions features, as well as the length of rows in the feature dictionary, are (32, 64, 128, 256). The number of rows in the dictionary are (512, 256, 128, 64), and the length of the read/write tags is 32 for all dictionaries.

\section{Training Strategy}

Although our FD-GAN implementation operates on 256$\times$256 images, the LD-Net part should in principle be independent of image size. For LD-Net we normalize the landmark coordinates such that the square bounding box of all points span the range [-1, 1]. For FD-GAN, pixel values are normalized to [-1, 1].

The training configuration is given in table \ref{tab:training}. Training time is in number of iterations.

\begin{table}
    \centering
    \begin{tabular}{ccccc}\hline
        Stage & Algorithm & LR & Batch & Time \\\hline
        LD-Net Stage 1 & Adam & $10^{-4}$ & 32 & $4\times 10^5$\\
        LD-Net Stage 2 & Adam & $5\times 10^{-5}$ & 32 & $10^6$\\
        FD-GAN & RMSprop & $2\times 10^{-5}$ & 4 & $10^6$\\\hline
    \end{tabular}
    \caption{Training configuration of both sub-networks.}
    \label{tab:training}
\end{table}
%\vspace{-0.5in}

\section{Performance}

For one target image and its corresponding landmarks, the identity code in LD-Net and the feature dictionary in the FD-GAN can be reused for multiple source images. For each target, we measure the running time using all 100 frames in a corresponding test video. %with one target frame per 100 source frames, since for our evaluations all test video clips are 100 frames long. 
Landmark detection is performed separately in advance and is not included in the running time. It takes approximately 0.08s for FD-GAN to generate one image and 0.02s for LD-Net to do landmark disentangling on a single NVIDIA TITANX GPU.

\section{Additional Qualitative Results}

\subsection{Ablation study}
\paragraph{Comparison between with and without LD-Net.}

\begin{figure*}[htp]
  \centering
  \setlength{\tabcolsep}{0.2mm}{
\begin{tabular}{cccccc}
  \includegraphics[width=0.16\linewidth]{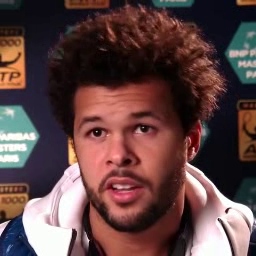} &
  \includegraphics[width=0.16\linewidth]{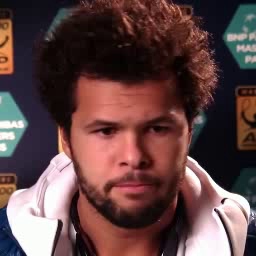} &
  \includegraphics[width=0.16\linewidth]{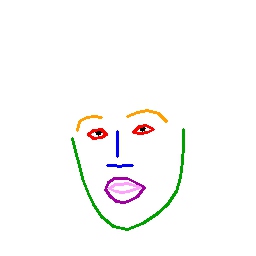} &
  \includegraphics[width=0.16\linewidth]{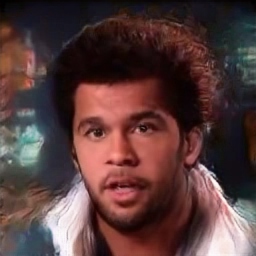} &
  \includegraphics[width=0.16\linewidth]{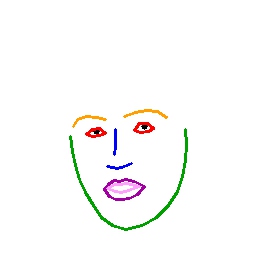} &
  \includegraphics[width=0.16\linewidth]{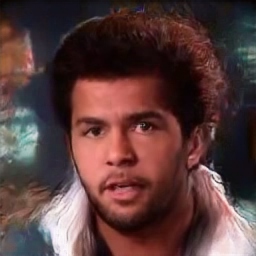} \\
  \includegraphics[width=0.16\linewidth]{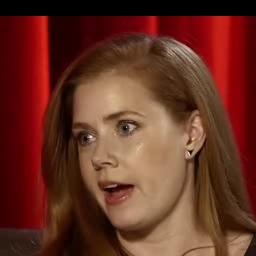} &
  \includegraphics[width=0.16\linewidth]{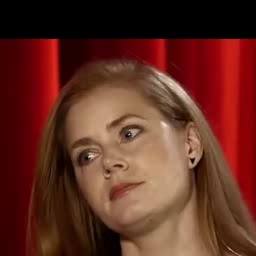} &
  \includegraphics[width=0.16\linewidth]{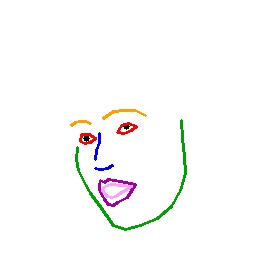} &
  \includegraphics[width=0.16\linewidth]{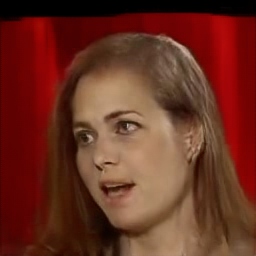} &
  \includegraphics[width=0.16\linewidth]{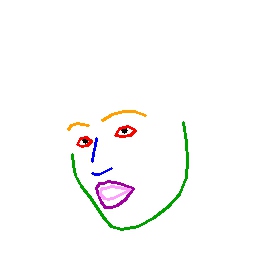} &
  \includegraphics[width=0.16\linewidth]{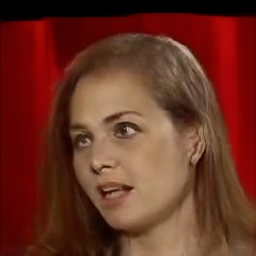} \\
  \includegraphics[width=0.16\linewidth]{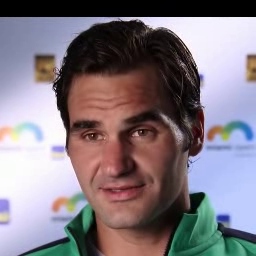} &
  \includegraphics[width=0.16\linewidth]{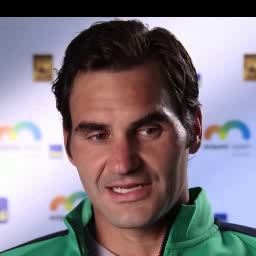} &
  \includegraphics[width=0.16\linewidth]{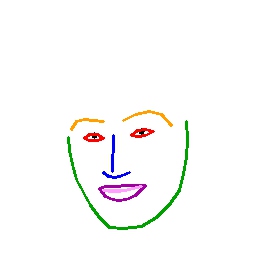} &
  \includegraphics[width=0.16\linewidth]{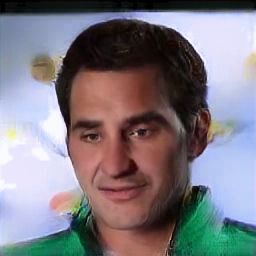} &
  \includegraphics[width=0.16\linewidth]{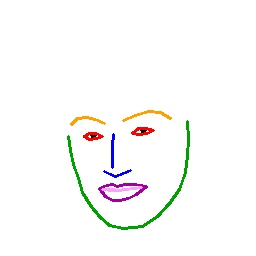} &
  \includegraphics[width=0.16\linewidth]{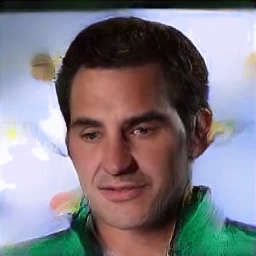} \\
  \includegraphics[width=0.16\linewidth]{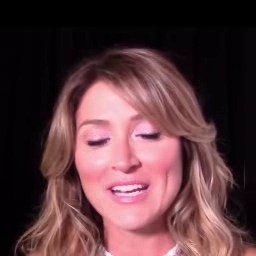} &
  \includegraphics[width=0.16\linewidth]{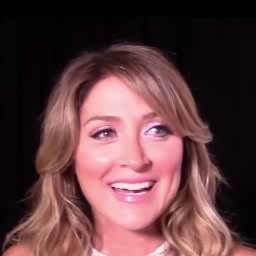} &
  \includegraphics[width=0.16\linewidth]{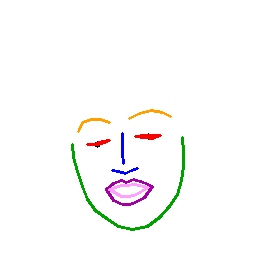} &
  \includegraphics[width=0.16\linewidth]{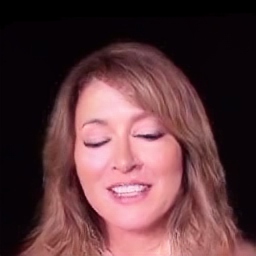} &
  \includegraphics[width=0.16\linewidth]{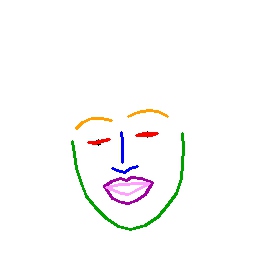} &
  \includegraphics[width=0.16\linewidth]{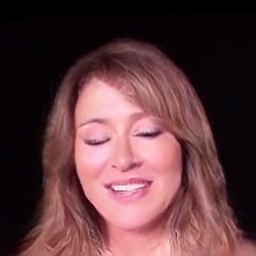} \\
  \includegraphics[width=0.16\linewidth]{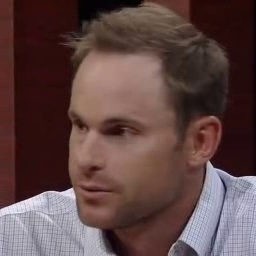} &
  \includegraphics[width=0.16\linewidth]{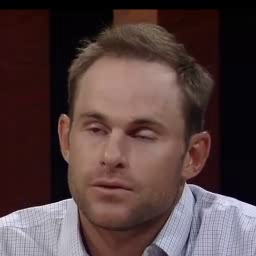} &
  \includegraphics[width=0.16\linewidth]{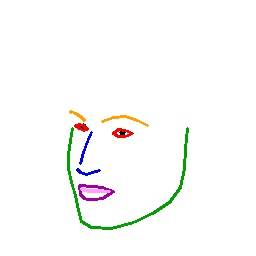} &
  \includegraphics[width=0.16\linewidth]{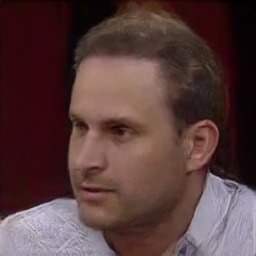} &
  \includegraphics[width=0.16\linewidth]{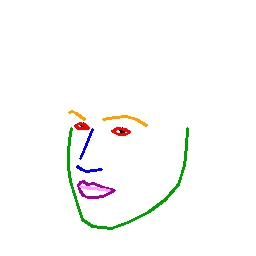} &
  \includegraphics[width=0.16\linewidth]{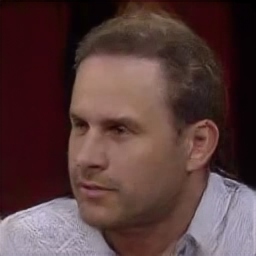} \\
  \includegraphics[width=0.16\linewidth]{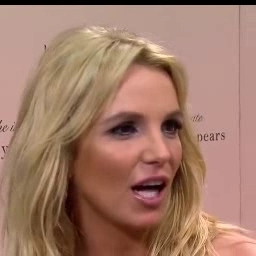} &
  \includegraphics[width=0.16\linewidth]{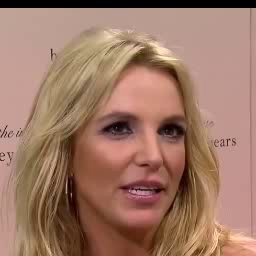} &
  \includegraphics[width=0.16\linewidth]{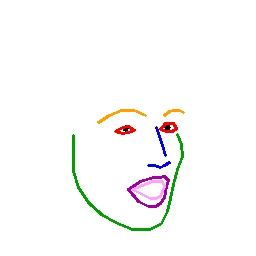} &
  \includegraphics[width=0.16\linewidth]{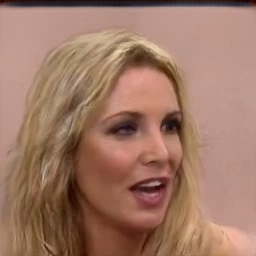} &
  \includegraphics[width=0.16\linewidth]{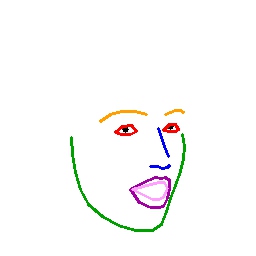} &
  \includegraphics[width=0.16\linewidth]{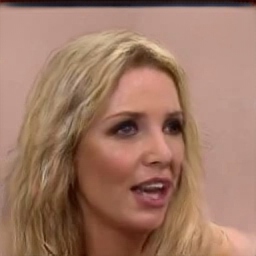} \\
  Source & Target & Ground Truth LMs & Result1 & LD-Net LMs & Result2
  \end{tabular}}
\caption{Qualitative comparison on self-reenactment between the results using ground truth landmarks and the results with synthetic landmarks by LD-Net (LMs is short for Landmarks).}
	\label{fig:comp1_1}
\end{figure*}

\begin{figure*}[htp]
  \centering
  \setlength{\tabcolsep}{0.2mm}{
\begin{tabular}{cccccc}
  \includegraphics[width=0.16\linewidth]{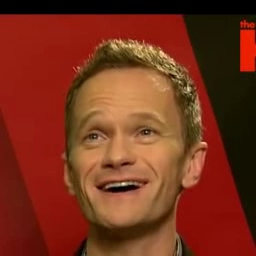} &
  \includegraphics[width=0.16\linewidth]{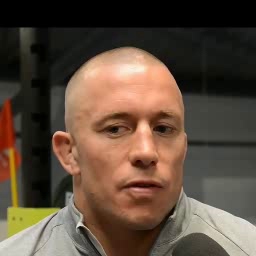} &
  \includegraphics[width=0.16\linewidth]{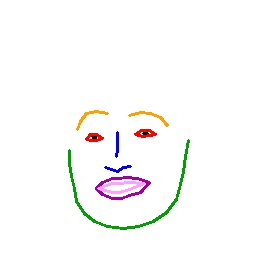} &
  \includegraphics[width=0.16\linewidth]{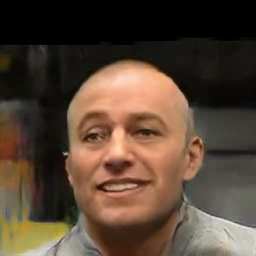} &
  \includegraphics[width=0.16\linewidth]{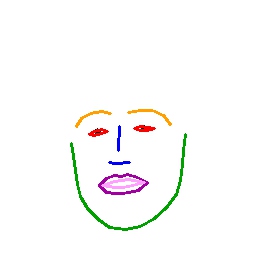} &
  \includegraphics[width=0.16\linewidth]{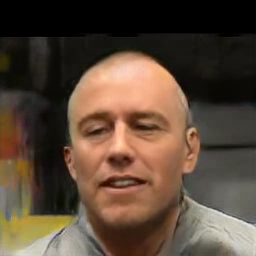} \\
  \includegraphics[width=0.16\linewidth]{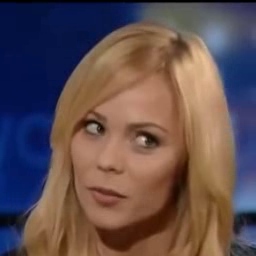} &
  \includegraphics[width=0.16\linewidth]{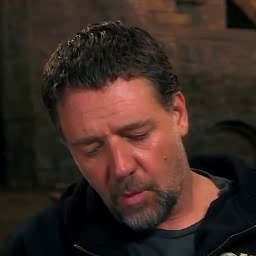} &
  \includegraphics[width=0.16\linewidth]{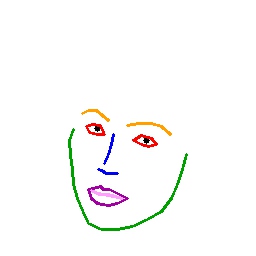} &
  \includegraphics[width=0.16\linewidth]{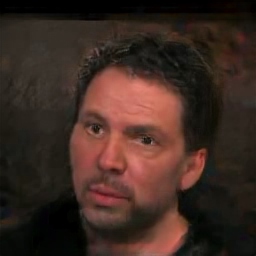} &
  \includegraphics[width=0.16\linewidth]{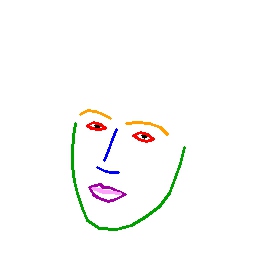} &
  \includegraphics[width=0.16\linewidth]{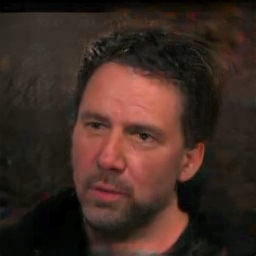} \\
  \includegraphics[width=0.16\linewidth]{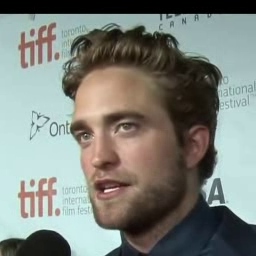} &
  \includegraphics[width=0.16\linewidth]{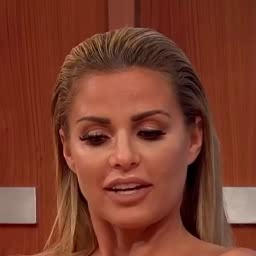} &
  \includegraphics[width=0.16\linewidth]{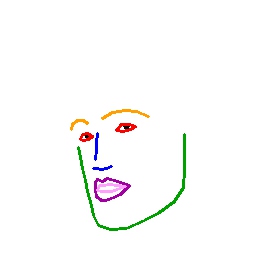} &
  \includegraphics[width=0.16\linewidth]{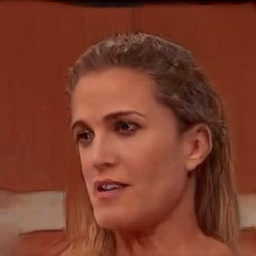} &
  \includegraphics[width=0.16\linewidth]{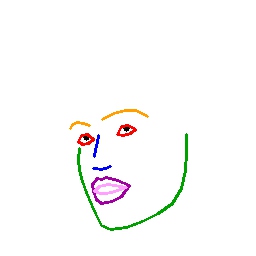} &
  \includegraphics[width=0.16\linewidth]{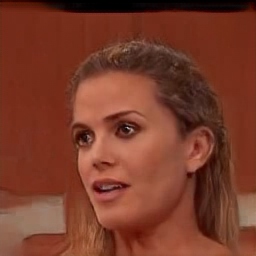} \\
  \includegraphics[width=0.16\linewidth]{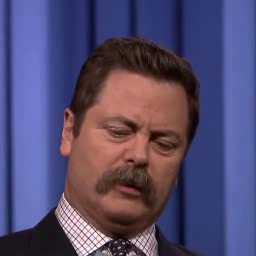} &
  \includegraphics[width=0.16\linewidth]{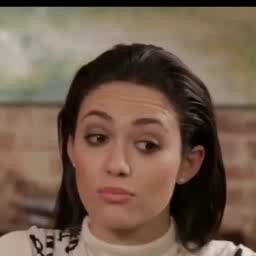} &
  \includegraphics[width=0.16\linewidth]{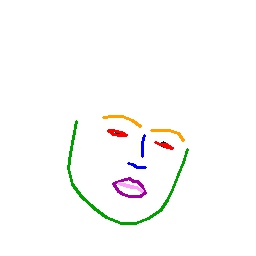} &
  \includegraphics[width=0.16\linewidth]{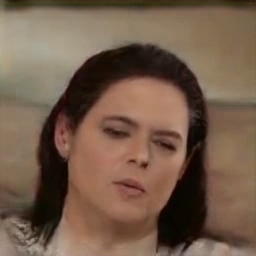} &
  \includegraphics[width=0.16\linewidth]{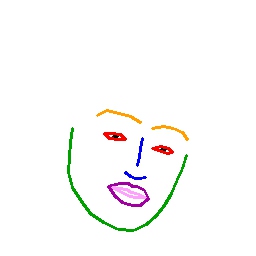} &
  \includegraphics[width=0.16\linewidth]{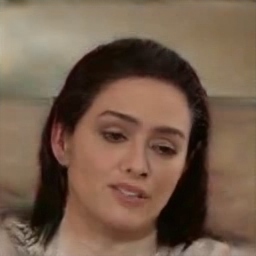} \\
  \includegraphics[width=0.16\linewidth]{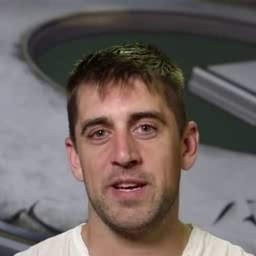} &
  \includegraphics[width=0.16\linewidth]{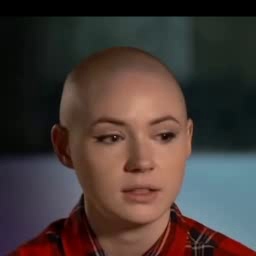} &
  \includegraphics[width=0.16\linewidth]{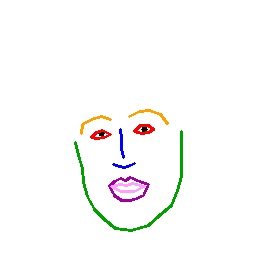} &
  \includegraphics[width=0.16\linewidth]{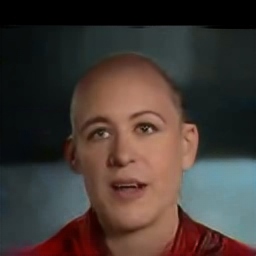} &
  \includegraphics[width=0.16\linewidth]{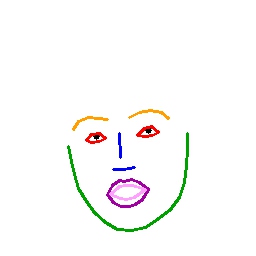} &
  \includegraphics[width=0.16\linewidth]{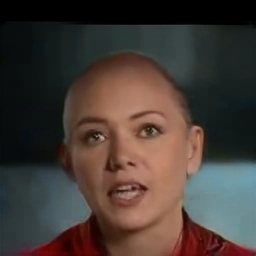} \\
  \includegraphics[width=0.16\linewidth]{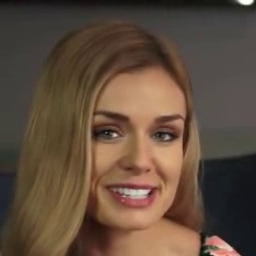} &
  \includegraphics[width=0.16\linewidth]{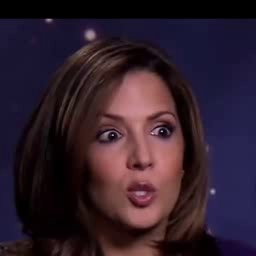} &
  \includegraphics[width=0.16\linewidth]{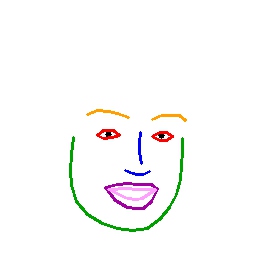} &
  \includegraphics[width=0.16\linewidth]{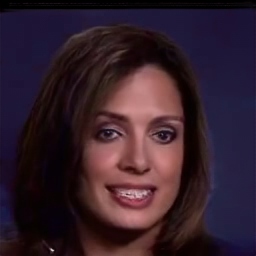} &
  \includegraphics[width=0.16\linewidth]{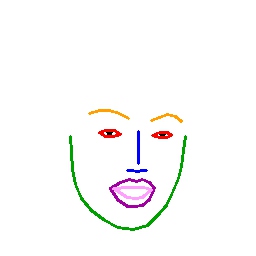} &
  \includegraphics[width=0.16\linewidth]{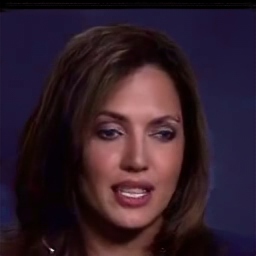} \\
  Source & Target & Source LMs & Result1 & LD-Net LMs & Result2
  \end{tabular}}
\caption{Qualitative comparison on cross-subject face reenactment between the results using source landmarks and the results with synthetic landmarks by LD-Net.}
	\label{fig:comp1_2}
\end{figure*}

In Fig.~\ref{fig:comp1_1} and Fig.~\ref{fig:comp1_2}, we show additional qualitative evaluations for self-reenactment and cross-subject face reenactment using the SelfTest dataset and CrossTest dataset, respectively. 
%Additional results for self-reenactment and cross-subject face reenactment on SelfTest dataset and CrossTest dataset are shown in Fig.~\ref{fig:comp1_1} and Fig.~\ref{fig:comp1_2} respectively. 
In Fig.~\ref{fig:comp1_1}, we compare results generated using ground truth landmarks (from the source video) and results using landmarks generated by LD-Net. 
As can be seen in the figure, our method can predict landmarks and synthesize high-quality images that are both close to the ground truth.
%In self-reenactment (Fig.~\ref{fig:comp1_1}), by comparing the results using source landmarks (ground truth landmarks) and the results using synthetic landmarks generated by LD-Net, 
%we show that our method is able to generate high-quality landmarks as well as images which are very close to the ground truth. 
For the cross-subject evaluation in Fig.~\ref{fig:comp1_2}, our results using landmarks by LD-Net not only have better identity preservation but also more precise poses/expressions (\textit{e.g.} in the first row).

\begin{figure*}[htp]
  \centering
  \setlength{\tabcolsep}{0.2mm}{
\begin{tabular}{cccccc}
  \includegraphics[width=0.16\linewidth]{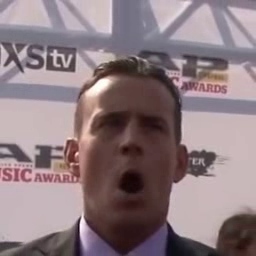} &
  \includegraphics[width=0.16\linewidth]{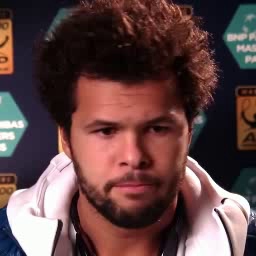} &
  \includegraphics[width=0.16\linewidth]{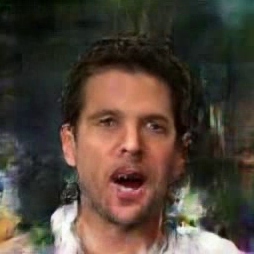} &
  \includegraphics[width=0.16\linewidth]{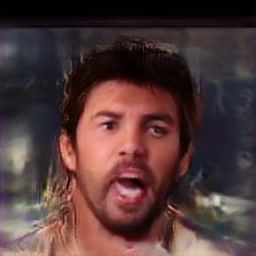} &
  \includegraphics[width=0.16\linewidth]{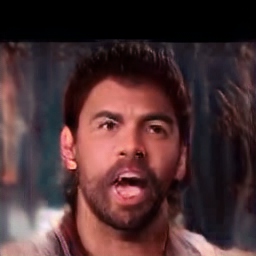} &
  \includegraphics[width=0.16\linewidth]{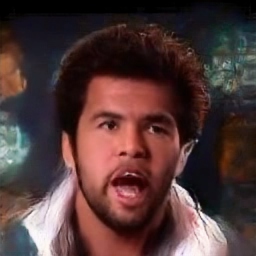} \\
  \includegraphics[width=0.16\linewidth]{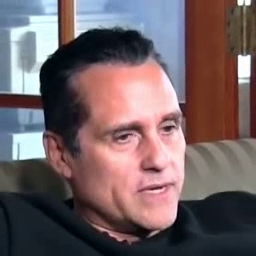} &
  \includegraphics[width=0.16\linewidth]{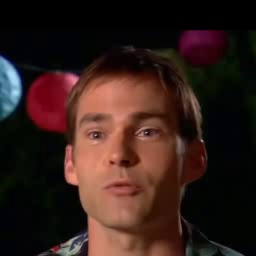} &
  \includegraphics[width=0.16\linewidth]{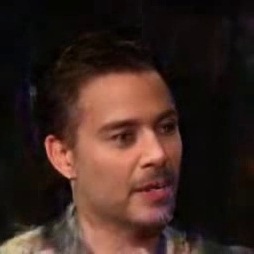} &
  \includegraphics[width=0.16\linewidth]{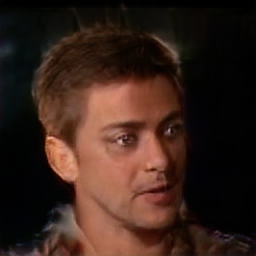} &
  \includegraphics[width=0.16\linewidth]{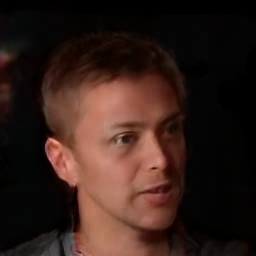} &
  \includegraphics[width=0.16\linewidth]{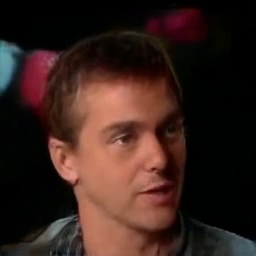} \\
  \includegraphics[width=0.16\linewidth]{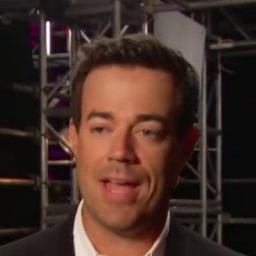} &
  \includegraphics[width=0.16\linewidth]{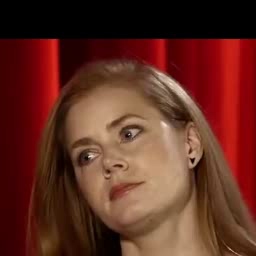} &
  \includegraphics[width=0.16\linewidth]{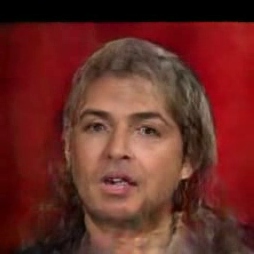} &
  \includegraphics[width=0.16\linewidth]{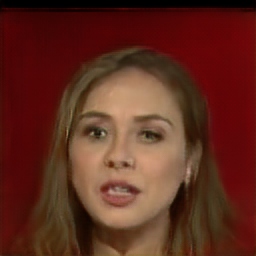} &
  \includegraphics[width=0.16\linewidth]{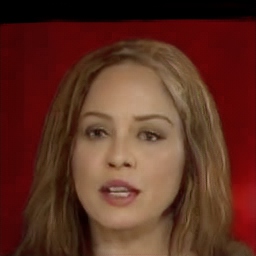} &
  \includegraphics[width=0.16\linewidth]{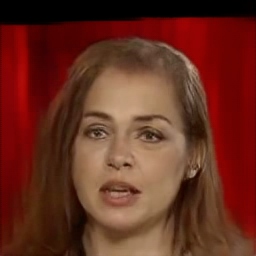} \\
  \includegraphics[width=0.16\linewidth]{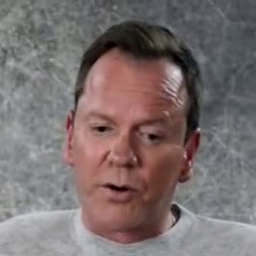} &
  \includegraphics[width=0.16\linewidth]{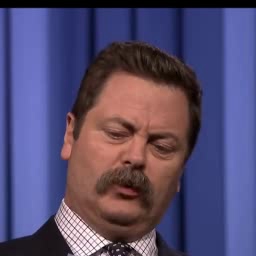} &
  \includegraphics[width=0.16\linewidth]{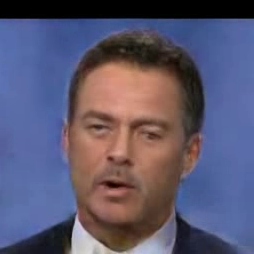} &
  \includegraphics[width=0.16\linewidth]{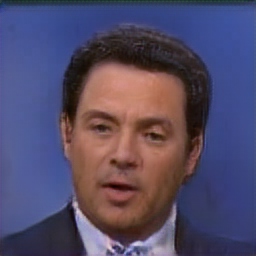} &
  \includegraphics[width=0.16\linewidth]{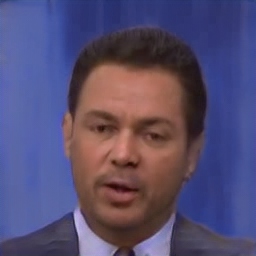} &
  \includegraphics[width=0.16\linewidth]{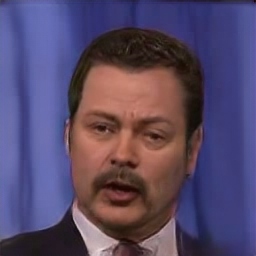} \\
  \includegraphics[width=0.16\linewidth]{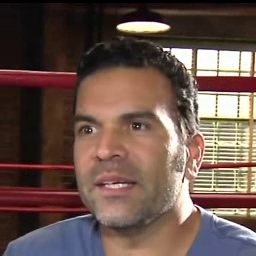} &
  \includegraphics[width=0.16\linewidth]{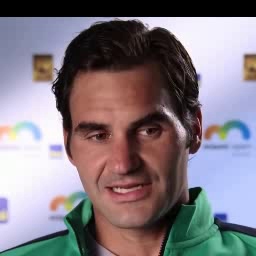} &
  \includegraphics[width=0.16\linewidth]{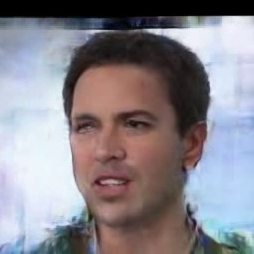} &
  \includegraphics[width=0.16\linewidth]{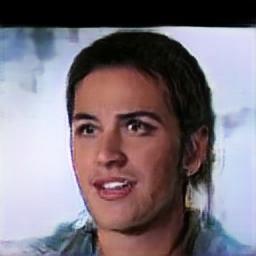} &
  \includegraphics[width=0.16\linewidth]{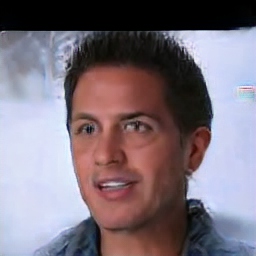} &
  \includegraphics[width=0.16\linewidth]{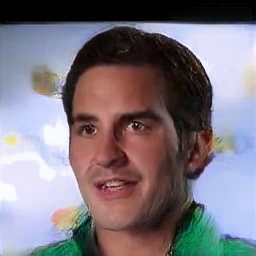} \\
  \includegraphics[width=0.16\linewidth]{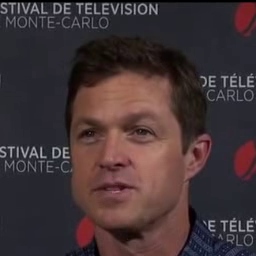} &
  \includegraphics[width=0.16\linewidth]{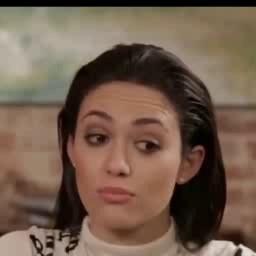} &
  \includegraphics[width=0.16\linewidth]{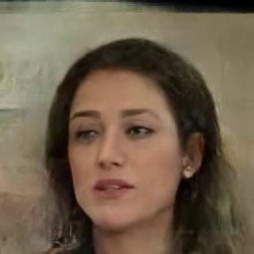} &
  \includegraphics[width=0.16\linewidth]{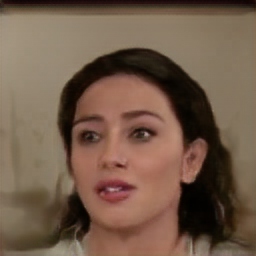} &
  \includegraphics[width=0.16\linewidth]{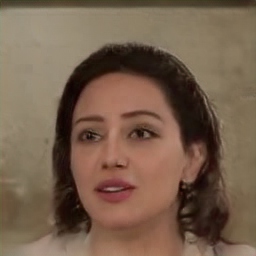} &
  \includegraphics[width=0.16\linewidth]{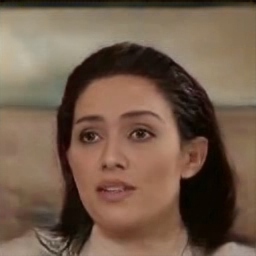} \\
  \includegraphics[width=0.16\linewidth]{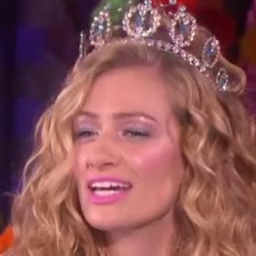} &
  \includegraphics[width=0.16\linewidth]{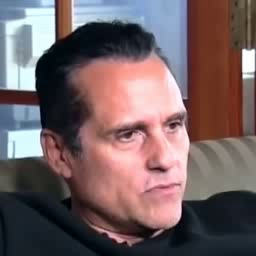} &
  \includegraphics[width=0.16\linewidth]{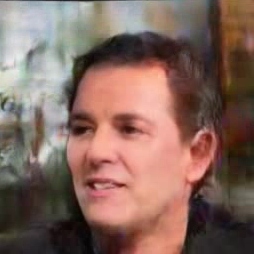} &
  \includegraphics[width=0.16\linewidth]{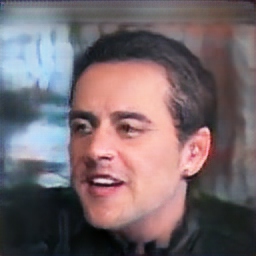} &
  \includegraphics[width=0.16\linewidth]{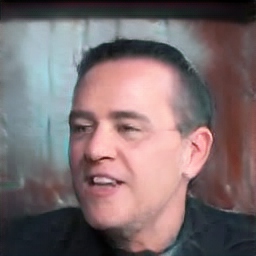} &
  \includegraphics[width=0.16\linewidth]{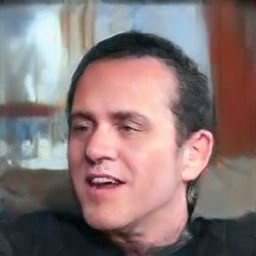} \\
  \includegraphics[width=0.16\linewidth]{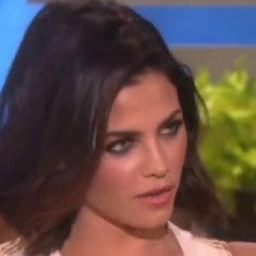} &
  \includegraphics[width=0.16\linewidth]{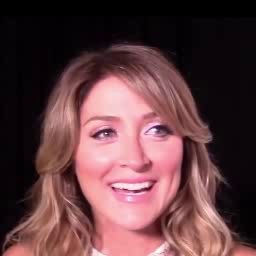} &
  \includegraphics[width=0.16\linewidth]{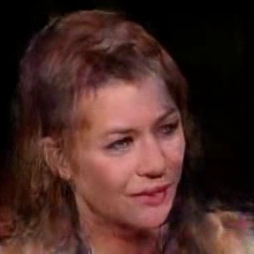} &
  \includegraphics[width=0.16\linewidth]{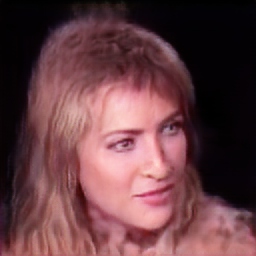} &
  \includegraphics[width=0.16\linewidth]{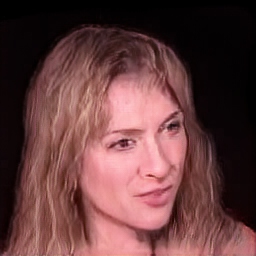} &
  \includegraphics[width=0.16\linewidth]{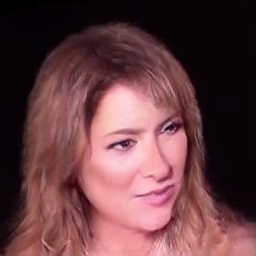} \\
  Source & Target & Pix2PixHD & AdaIN & FD-GAN-1 & Ours \\
  \end{tabular}}
\caption{Qualitative comparison on cross-subject face reenactment between our FD-GAN and the baselines: Pix2PixHD~\cite{wang2018pix2pixHD}, AdaIN and FD-GAN-1.}
	\label{fig:comp2}
\end{figure*}

\paragraph{Comparison FD-GAN with baselines.}
Fig.~\ref{fig:comp2} shows additional qualitative comparisons for cross-subject face reenactment on the CrossTest dataset with the three baselines, including the advanced image-to-image translation network Pix-2PixHD~\cite{wang2018pix2pixHD}, as well as two variants of FD-GAN (AdaIN and FD-GAN-1). AdaIN (colum 4) and FD-GAN-1 (column 5) use the full feature dictionary with AdaIN~\cite{huang2017adain} or a one-line feature dictionary, respectively.

\subsection{Comparison with one-shot methods}
\begin{figure*}[htp]
  \centering
  \setlength{\tabcolsep}{0.2mm}{
\begin{tabular}{cccccc}
  
  \includegraphics[width=0.16\linewidth]{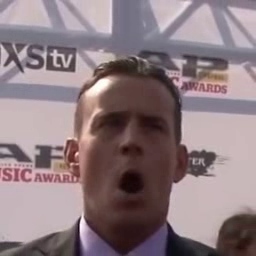} &
  \includegraphics[width=0.16\linewidth]{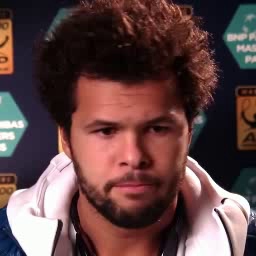} &
  \includegraphics[width=0.16\linewidth]{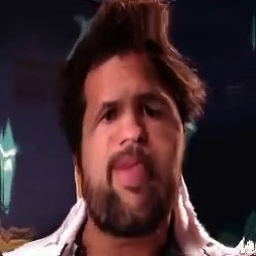} &
  \includegraphics[width=0.16\linewidth]{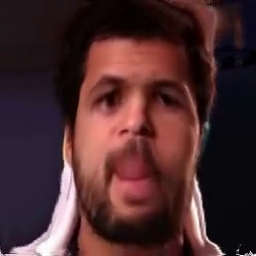} &
  \includegraphics[width=0.16\linewidth]{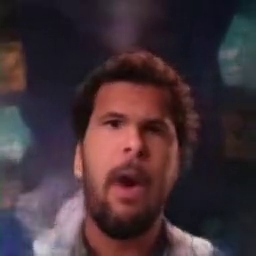} &
  \includegraphics[width=0.16\linewidth]{figure/fig2/fig2_fd/00_05.jpg}\\
  
  \includegraphics[width=0.16\linewidth]{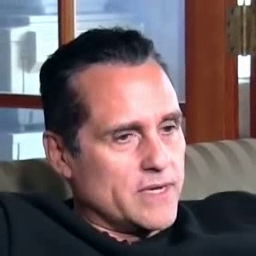} &
  \includegraphics[width=0.16\linewidth]{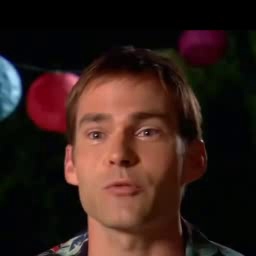} &
  \includegraphics[width=0.16\linewidth]{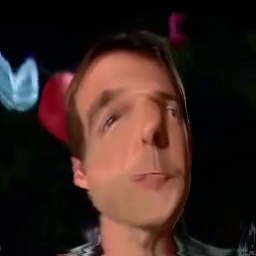} &
  \includegraphics[width=0.16\linewidth]{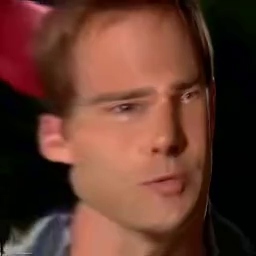} &
  \includegraphics[width=0.16\linewidth]{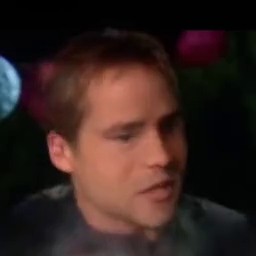} &
  \includegraphics[width=0.16\linewidth]{figure/fig2/fig2_fd/01_05.jpg} \\
  
  \includegraphics[width=0.16\linewidth]{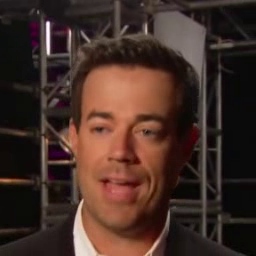} &
  \includegraphics[width=0.16\linewidth]{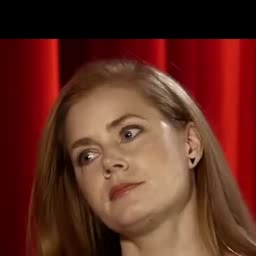} &
  \includegraphics[width=0.16\linewidth]{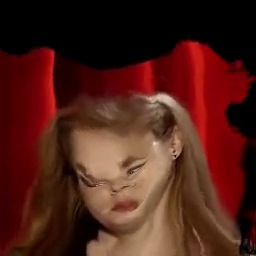} &
  \includegraphics[width=0.16\linewidth]{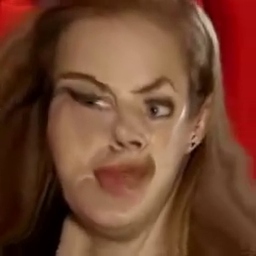} &
  \includegraphics[width=0.16\linewidth]{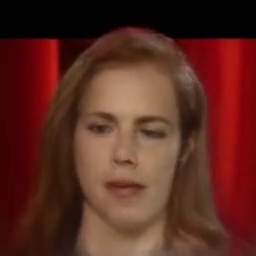} &
  \includegraphics[width=0.16\linewidth]{figure/fig2/fig2_fd/02_05.jpg} \\
  
  \includegraphics[width=0.16\linewidth]{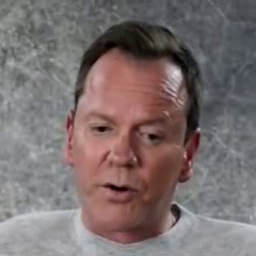} &
  \includegraphics[width=0.16\linewidth]{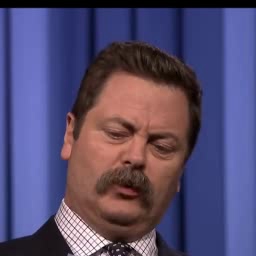} &
  \includegraphics[width=0.16\linewidth]{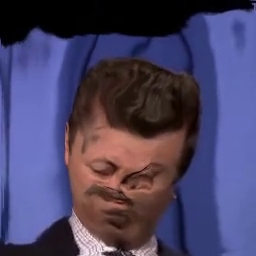} &
  \includegraphics[width=0.16\linewidth]{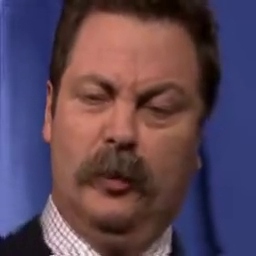} &
  \includegraphics[width=0.16\linewidth]{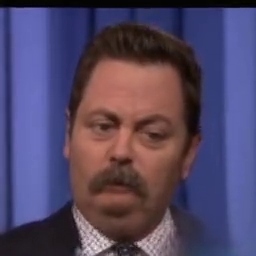} &
  \includegraphics[width=0.16\linewidth]{figure/fig2/fig2_fd/03_05.jpg}\\
  
  \includegraphics[width=0.16\linewidth]{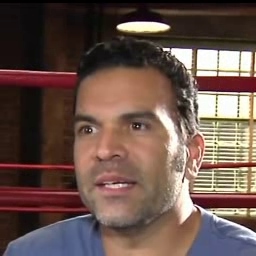} &
  \includegraphics[width=0.16\linewidth]{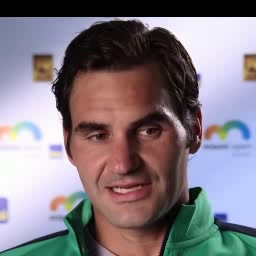} &
  \includegraphics[width=0.16\linewidth]{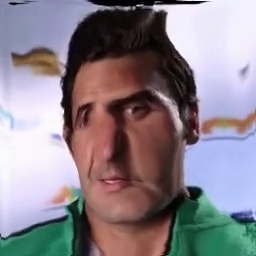} &
  \includegraphics[width=0.16\linewidth]{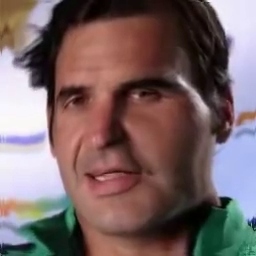} &
  \includegraphics[width=0.16\linewidth]{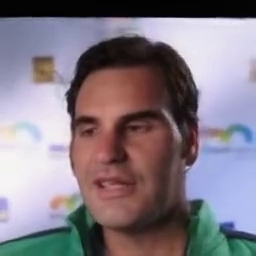} &
  \includegraphics[width=0.16\linewidth]{figure/fig2/fig2_fd/04_05.jpg} \\
  
  \includegraphics[width=0.16\linewidth]{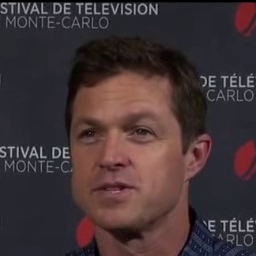} &
  \includegraphics[width=0.16\linewidth]{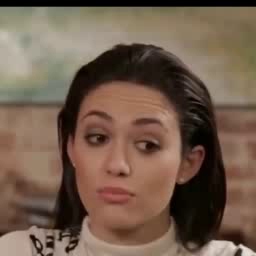} &
  \includegraphics[width=0.16\linewidth]{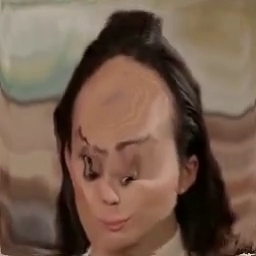} &
  \includegraphics[width=0.16\linewidth]{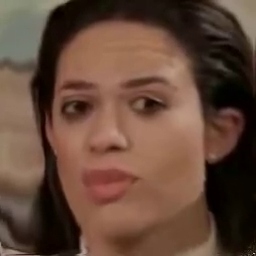} &
  \includegraphics[width=0.16\linewidth]{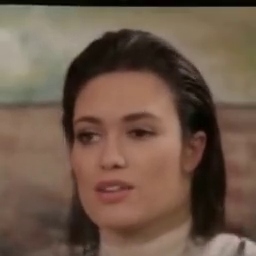} &
  \includegraphics[width=0.16\linewidth]{figure/fig2/fig2_fd/05_05.jpg} \\
  
  \includegraphics[width=0.16\linewidth]{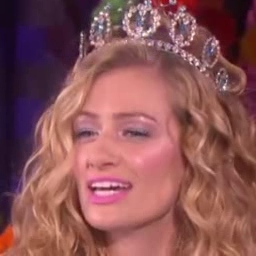} &
  \includegraphics[width=0.16\linewidth]{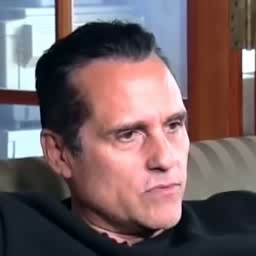} &
  \includegraphics[width=0.16\linewidth]{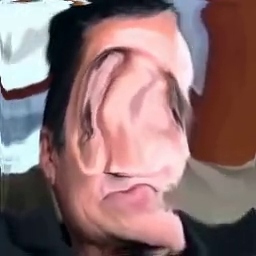} &
  \includegraphics[width=0.16\linewidth]{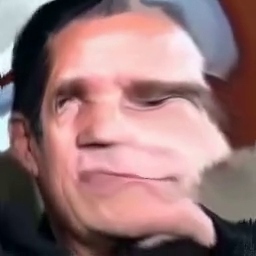} &
  \includegraphics[width=0.16\linewidth]{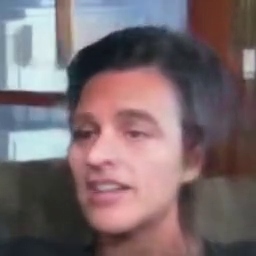} &
  \includegraphics[width=0.16\linewidth]{figure/fig2/fig2_fd/06_05.jpg} \\
  
  \includegraphics[width=0.16\linewidth]{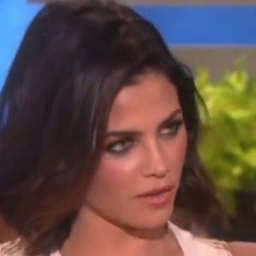} &
  \includegraphics[width=0.16\linewidth]{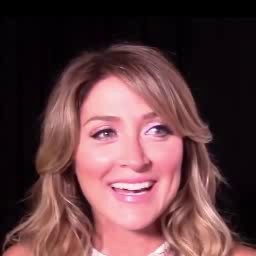} &
  \includegraphics[width=0.16\linewidth]{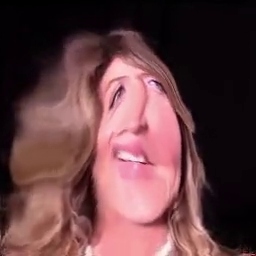} &
  \includegraphics[width=0.16\linewidth]{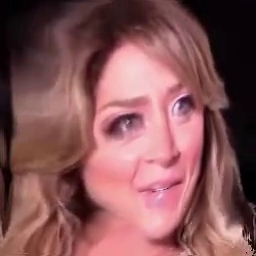} &
  \includegraphics[width=0.16\linewidth]{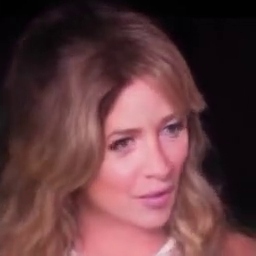} &
  \includegraphics[width=0.16\linewidth]{figure/fig2/fig2_fd/07_05.jpg} \\

  Source & Target & X2face & X2face & First-order & Ours \\
  & & & -aligned & -model & \\
  \end{tabular}}
  \caption{Qualitative comparison on cross-subject face reenactment using the CrossTest dataset between our method and one-shot methods: X2face~\cite{Wiles_2018_ECCV}, X2face-aligned~\cite{Wiles_2018_ECCV} and First-order-model~\cite{NIPS2019FirstOrder}.}
	\label{fig:comp3_1}
\end{figure*}

\begin{figure*}[htp]
  \centering
  \setlength{\tabcolsep}{0.2mm}{
  {
\begin{tabular}{cccccc}
    \includegraphics[width=0.16\linewidth]{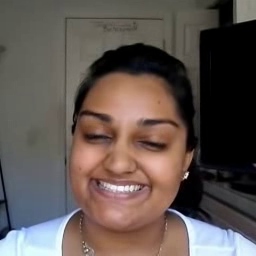} &
    \includegraphics[width=0.16\linewidth]{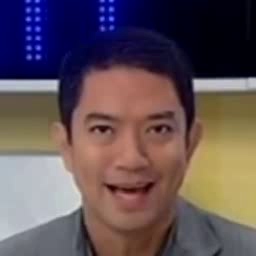} &
    \includegraphics[width=0.16\linewidth]{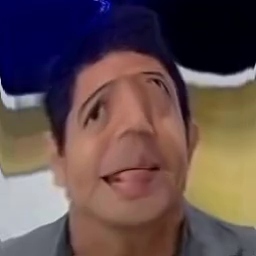} &
    \includegraphics[width=0.16\linewidth]{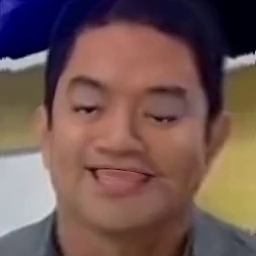} &
    \includegraphics[width=0.16\linewidth]{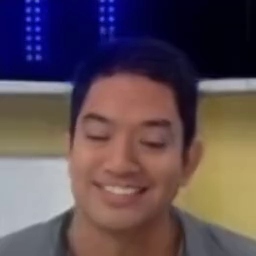} &
    \includegraphics[width=0.16\linewidth]{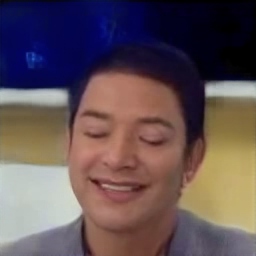} \\
    
    \includegraphics[width=0.16\linewidth]{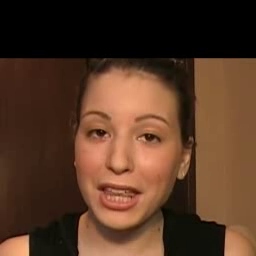} &
    \includegraphics[width=0.16\linewidth]{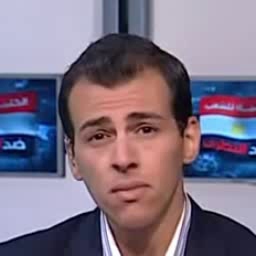} &
    \includegraphics[width=0.16\linewidth]{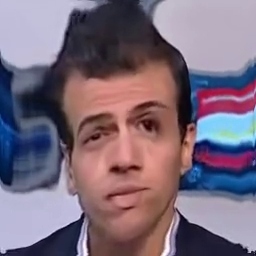} &
    \includegraphics[width=0.16\linewidth]{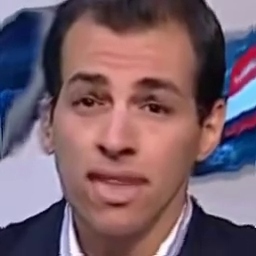} &
    \includegraphics[width=0.16\linewidth]{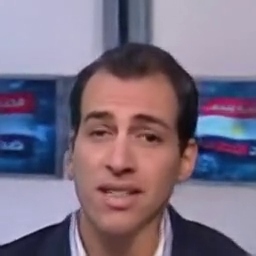} &
    \includegraphics[width=0.16\linewidth]{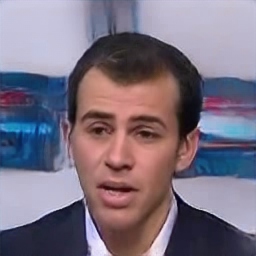} \\
    
    \includegraphics[width=0.16\linewidth]{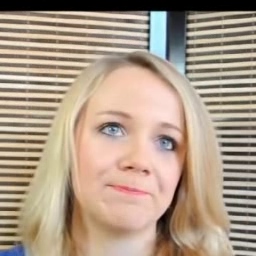} &
    \includegraphics[width=0.16\linewidth]{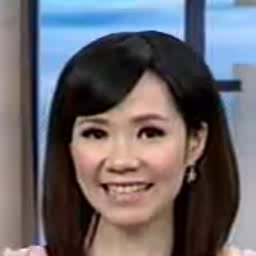} &
    \includegraphics[width=0.16\linewidth]{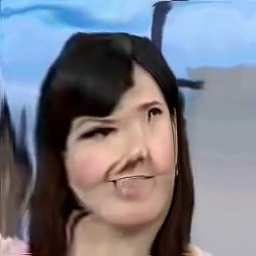} &
    \includegraphics[width=0.16\linewidth]{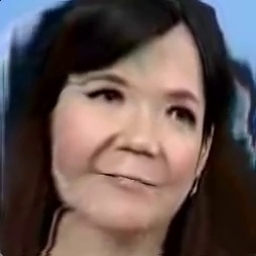} &
    \includegraphics[width=0.16\linewidth]{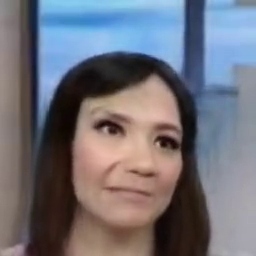} &
    \includegraphics[width=0.16\linewidth]{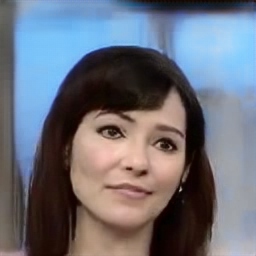} \\
    
    \includegraphics[width=0.16\linewidth]{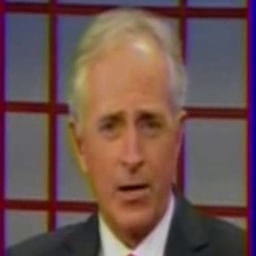} &
    \includegraphics[width=0.16\linewidth]{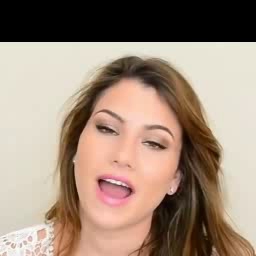} &
    \includegraphics[width=0.16\linewidth]{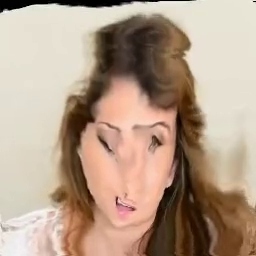} &
    \includegraphics[width=0.16\linewidth]{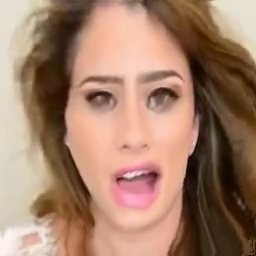} &
    \includegraphics[width=0.16\linewidth]{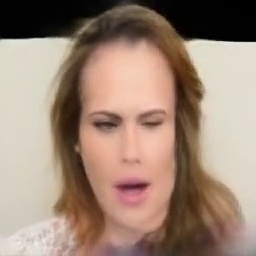} &
    \includegraphics[width=0.16\linewidth]{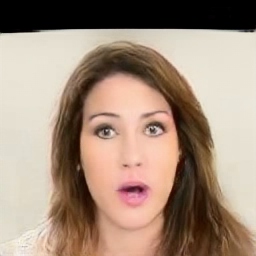} \\
    
    \includegraphics[width=0.16\linewidth]{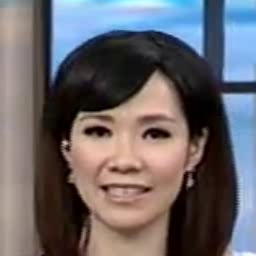} &
    \includegraphics[width=0.16\linewidth]{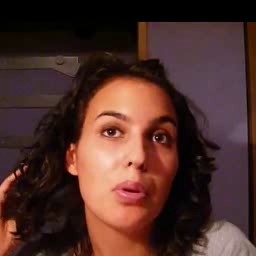} &
    \includegraphics[width=0.16\linewidth]{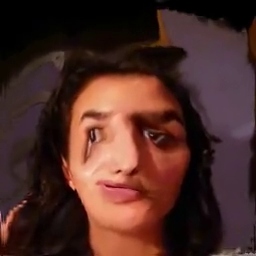} &
    \includegraphics[width=0.16\linewidth]{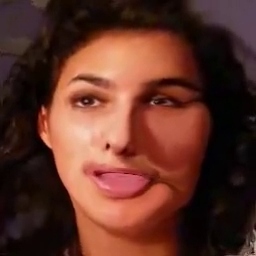} &
    \includegraphics[width=0.16\linewidth]{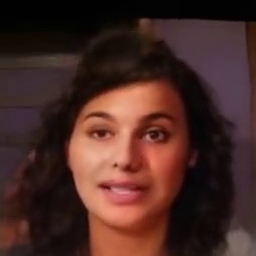} &
    \includegraphics[width=0.16\linewidth]{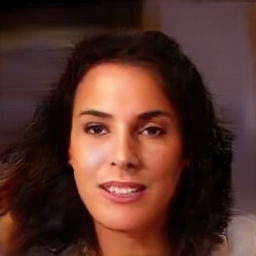} \\
    
    \includegraphics[width=0.16\linewidth]{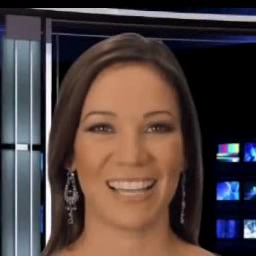} &
    \includegraphics[width=0.16\linewidth]{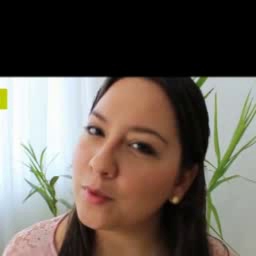} &
    \includegraphics[width=0.16\linewidth]{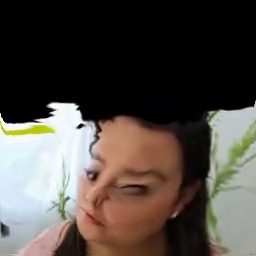} &
    \includegraphics[width=0.16\linewidth]{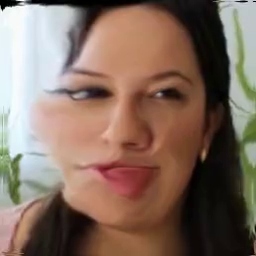} &
    \includegraphics[width=0.16\linewidth]{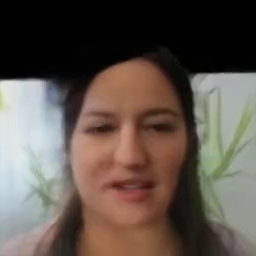} &
    \includegraphics[width=0.16\linewidth]{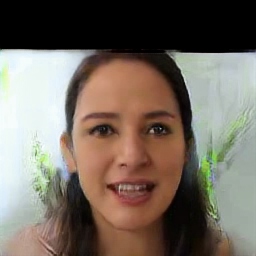} \\
    
    \includegraphics[width=0.16\linewidth]{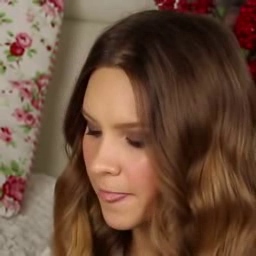} &
    \includegraphics[width=0.16\linewidth]{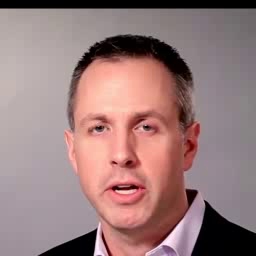} &
    \includegraphics[width=0.16\linewidth]{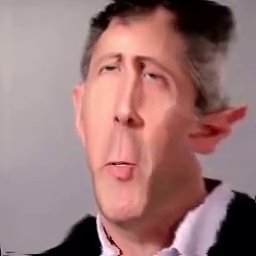} &
    \includegraphics[width=0.16\linewidth]{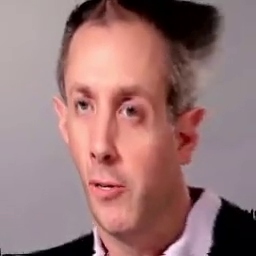} &
    \includegraphics[width=0.16\linewidth]{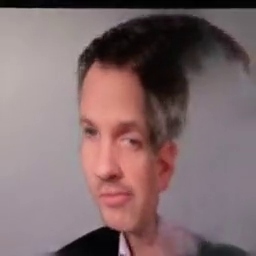} &
    \includegraphics[width=0.16\linewidth]{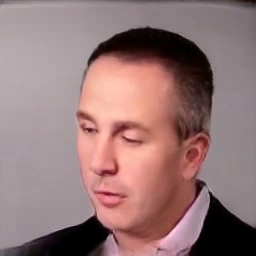} \\
    
    \includegraphics[width=0.16\linewidth]{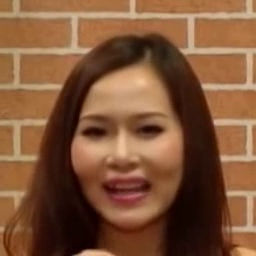} &
    \includegraphics[width=0.16\linewidth]{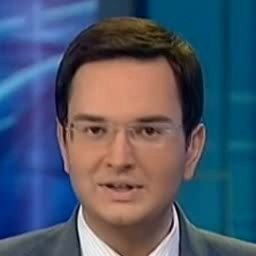} &
    \includegraphics[width=0.16\linewidth]{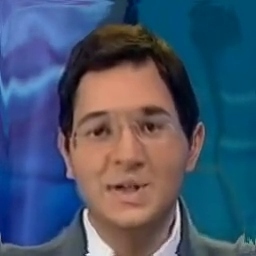} &
    \includegraphics[width=0.16\linewidth]{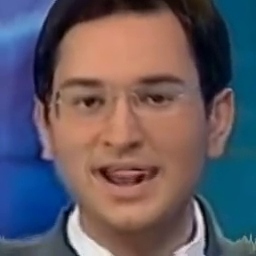} &
    \includegraphics[width=0.16\linewidth]{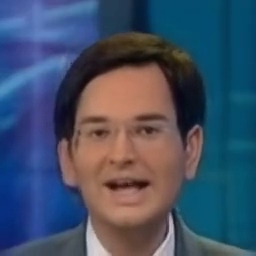} &
    \includegraphics[width=0.16\linewidth]{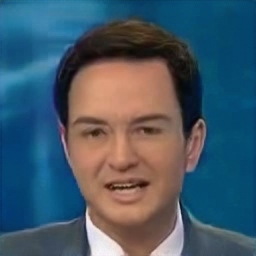} \\
    
    Source & Target & X2face & X2face-align & First-order & Ours \\
  & & & & -model & \\
  \end{tabular}}}
  \caption{Qualitative comparison on cross-subject face reenactment using FaceForensics++ dataset~\cite{roessler2019faceforensicspp} between our method and one-shot methods: X2face~\cite{Wiles_2018_ECCV}, X2face-aligned~\cite{Wiles_2018_ECCV} and First-order-model~\cite{NIPS2019FirstOrder}.}
	\label{fig:comp3_2}
\end{figure*}

Additional results for one-shot cross-subject face reenactment on the CrossTest dataset and FaceForensics++ dataset~\cite{roessler2019faceforensicspp} are shown in Fig.~\ref{fig:comp3_1} and Fig.~\ref{fig:comp3_2} respectively, with comparisons between our method and X2face~\cite{Wiles_2018_ECCV} (column 3), X2face-aligned~\cite{Wiles_2018_ECCV} (column 4), and First-order-model~\cite{NIPS2019FirstOrder} (column 5). Note that in X2face~\cite{Wiles_2018_ECCV}, the generated frames inherit the object proportions of the driving source video by transferring absolute coordinates, and thus it is very sensitive to face alignment. In addition to testing X2face~\cite{Wiles_2018_ECCV} using exactly the same input configurations as other methods, we also take a smaller face region with a tighter bounding box as input to minimize the misalignment between source and target images for X2face~\cite{Wiles_2018_ECCV} and obtain the results as shown as X2face-aligned. We also compute the metrics of the results by X2face-aligned on the CrossTest dataset, which are ISIM=0.7855, PSIM=0.7207, ED=0.344 and FID=62.14. Although the results are better than using non-aligned face images, our results are both quantitatively and qualitatively superior to theirs.

\subsection{Comparison with 3D-based methods}
\begin{figure*}[htp]
  \centering
  \setlength{\tabcolsep}{0.2mm}{
\begin{tabular}{ccccc}
  \includegraphics[width=0.168\linewidth]{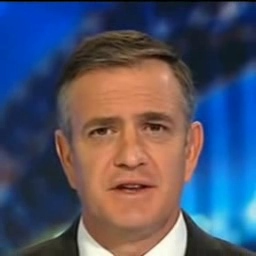} &
  \includegraphics[width=0.168\linewidth]{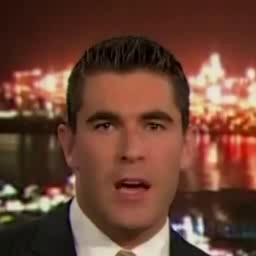} &
  \includegraphics[width=0.168\linewidth]{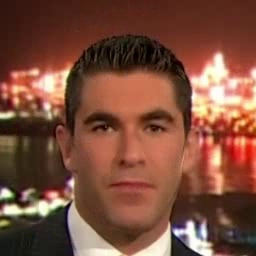} &
  \includegraphics[width=0.168\linewidth]{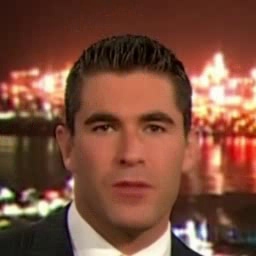} &
  \includegraphics[width=0.168\linewidth]{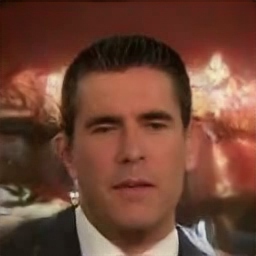} \\
  \includegraphics[width=0.168\linewidth]{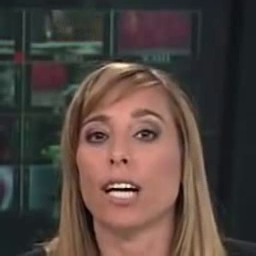} &
  \includegraphics[width=0.168\linewidth]{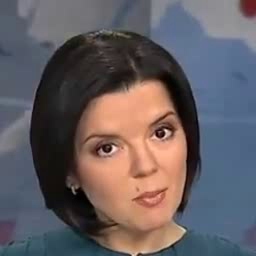} &
  \includegraphics[width=0.168\linewidth]{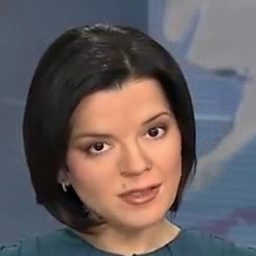} &
  \includegraphics[width=0.168\linewidth]{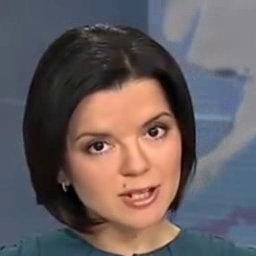} &
  \includegraphics[width=0.168\linewidth]{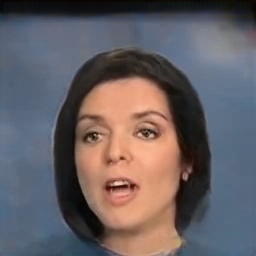} \\
  \includegraphics[width=0.168\linewidth]{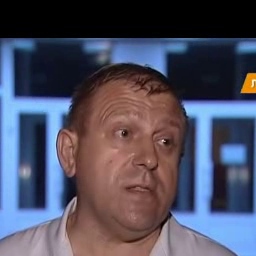} &
  \includegraphics[width=0.168\linewidth]{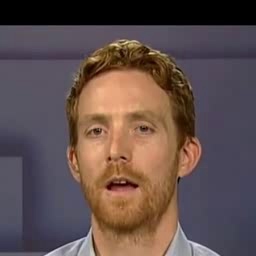} &
  \includegraphics[width=0.168\linewidth]{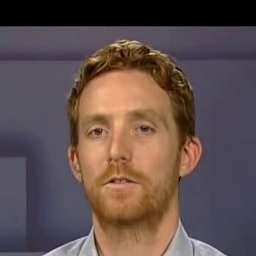} &
  \includegraphics[width=0.168\linewidth]{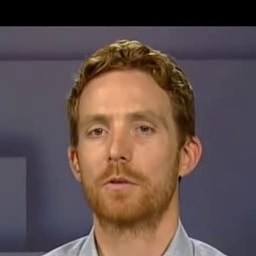} &
  \includegraphics[width=0.168\linewidth]{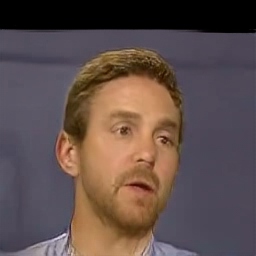} \\
  \includegraphics[width=0.168\linewidth]{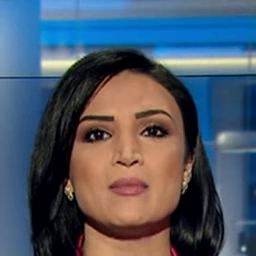} &
  \includegraphics[width=0.168\linewidth]{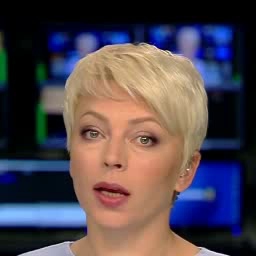} &
  \includegraphics[width=0.168\linewidth]{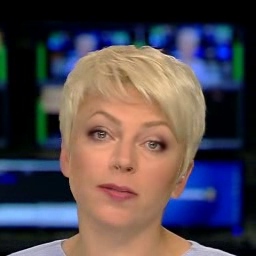} &
  \includegraphics[width=0.168\linewidth]{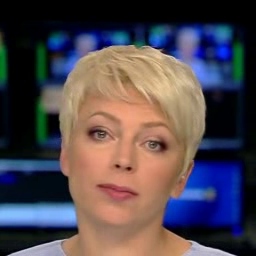} &
  \includegraphics[width=0.168\linewidth]{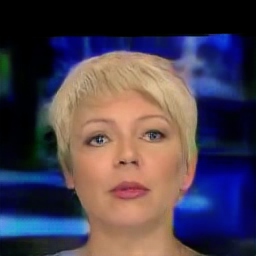} \\
  \includegraphics[width=0.168\linewidth]{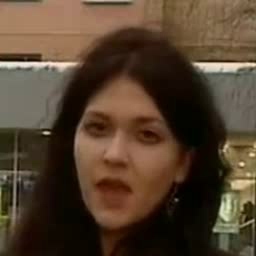} &
  \includegraphics[width=0.168\linewidth]{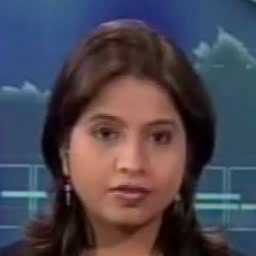} &
  \includegraphics[width=0.168\linewidth]{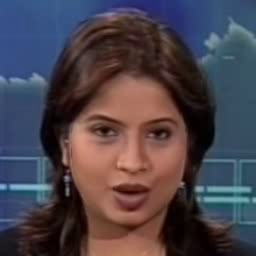} &
  \includegraphics[width=0.168\linewidth]{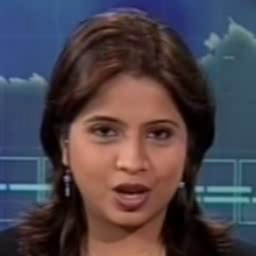} &
  \includegraphics[width=0.168\linewidth]{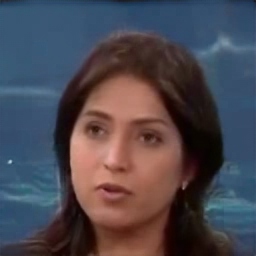} \\
  \includegraphics[width=0.168\linewidth]{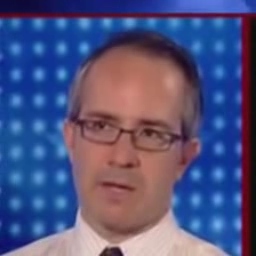} &
  \includegraphics[width=0.168\linewidth]{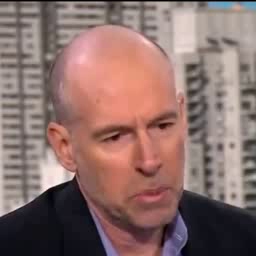} &
  \includegraphics[width=0.168\linewidth]{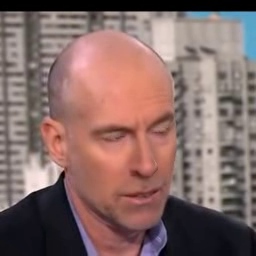} &
  \includegraphics[width=0.168\linewidth]{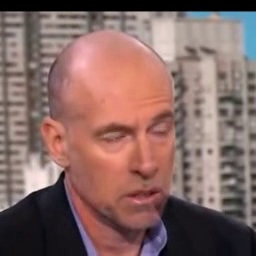} &
  \includegraphics[width=0.168\linewidth]{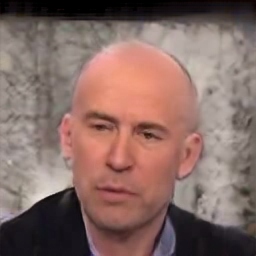} \\
  \includegraphics[width=0.168\linewidth]{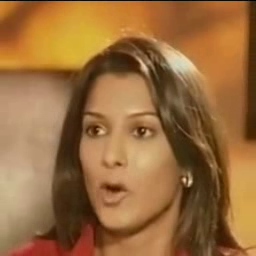} &
  \includegraphics[width=0.168\linewidth]{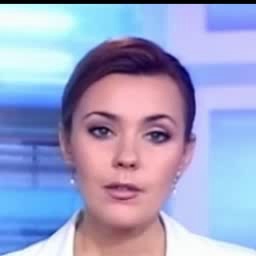} &
  \includegraphics[width=0.168\linewidth]{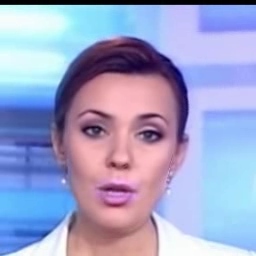} &
  \includegraphics[width=0.168\linewidth]{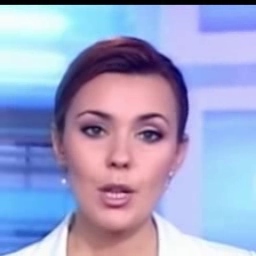} &
  \includegraphics[width=0.168\linewidth]{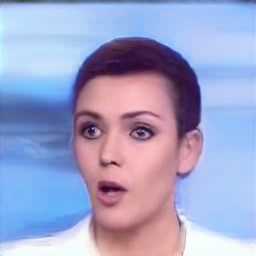} \\
  \includegraphics[width=0.168\linewidth]{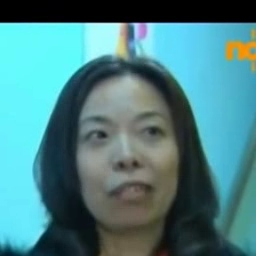} &
  \includegraphics[width=0.168\linewidth]{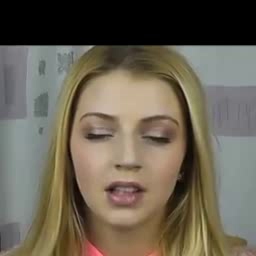} &
  \includegraphics[width=0.168\linewidth]{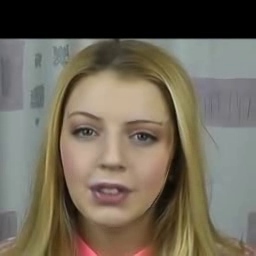} &
  \includegraphics[width=0.168\linewidth]{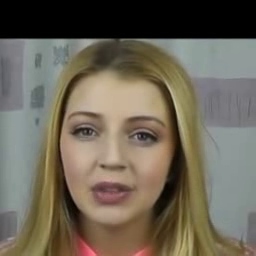} &
  \includegraphics[width=0.168\linewidth]{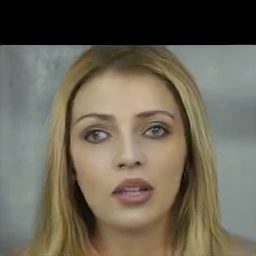} \\
  Source & Target & Face2face & NeuralTexture & Ours \\
  \end{tabular}}
  \caption{Qualitative comparison on cross-subject face reenactment with Face2face~\cite{thies2016face2face} and NeuralTexture~\cite{thies2019deferred} using input from FaceForensics++~\cite{roessler2019faceforensicspp}. Note that their methods control facial expressions only while ours can handle both the head pose and facial expressions.}
  \label{fig:comp5}
\end{figure*}

Additional results for cross-subject face reenactment on the FaceForensics++ dataset~\cite{roessler2019faceforensicspp} are shown in Fig.~\ref{fig:comp5}, with comparisons to two state-of-the-art 3D-based face reenactment methods, Face2Face~\cite{thies2016face2face} and NeuralTexture~\cite{thies2019deferred}.

% \begin{figure*}[t]
% 	\footnotesize
% 	\centering
% 	\includegraphics[width=1.0\linewidth]{figure/puppet_master1.jpg}
% 	\caption{Qualitative comparison on cross-subject face reenactment with Face2Face~\cite{thies2016face2face} and NeuralTexture~\cite{thies2019deferred} using input from FaceForensicss++~\cite{roessler2019faceforensicspp}. Note that their methods control facial expression only while ours can handle both head pose and facial expression.}
% 	\label{fig:comp5}
% \end{figure*}

\subsection{More Results}
\begin{figure*}[htp]
  \centering
  \setlength{\tabcolsep}{0.2mm}{
\begin{tabular}{cccc}
  \includegraphics[width=0.16\linewidth]{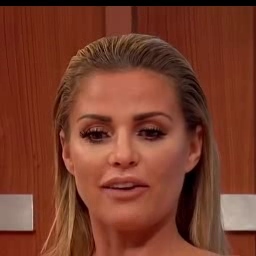} &
  \includegraphics[width=0.16\linewidth]{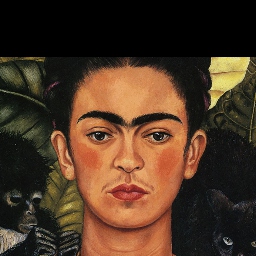} &
  \includegraphics[width=0.16\linewidth]{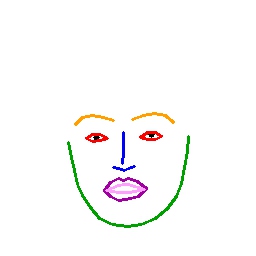} &
  \includegraphics[width=0.16\linewidth]{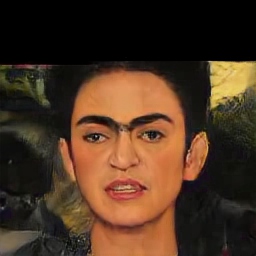} \\
  \includegraphics[width=0.16\linewidth]{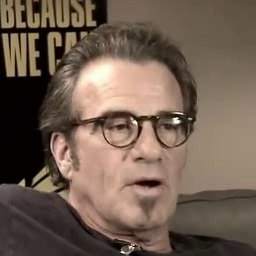} &
  \includegraphics[width=0.16\linewidth]{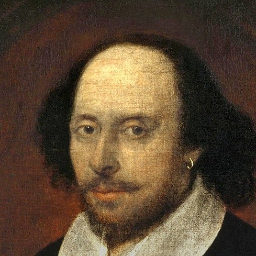} &
  \includegraphics[width=0.16\linewidth]{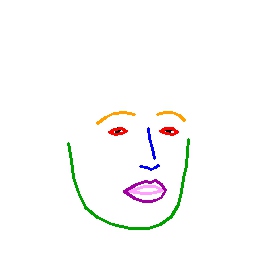} &
  \includegraphics[width=0.16\linewidth]{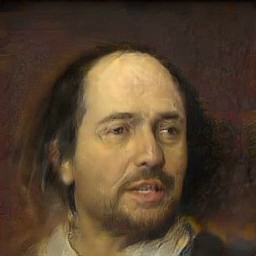} \\
  \includegraphics[width=0.16\linewidth]{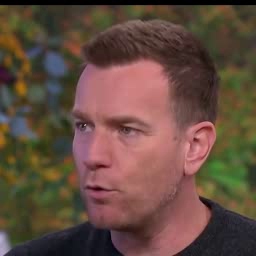} &
  \includegraphics[width=0.16\linewidth]{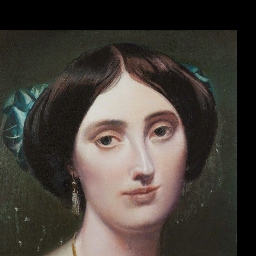} &
  \includegraphics[width=0.16\linewidth]{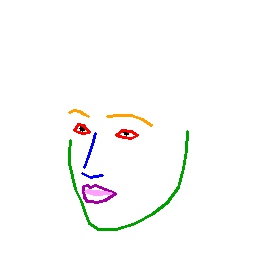} &
  \includegraphics[width=0.16\linewidth]{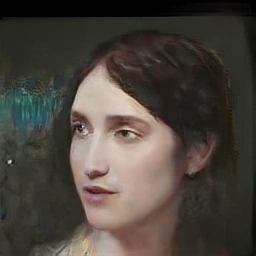} \\
  \includegraphics[width=0.16\linewidth]{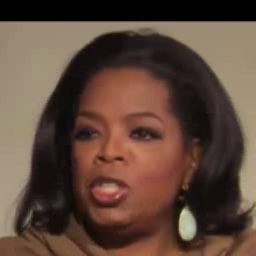} &
  \includegraphics[width=0.16\linewidth]{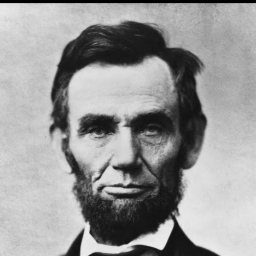} &
  \includegraphics[width=0.16\linewidth]{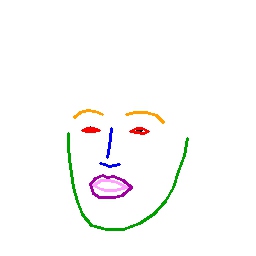} &
  \includegraphics[width=0.16\linewidth]{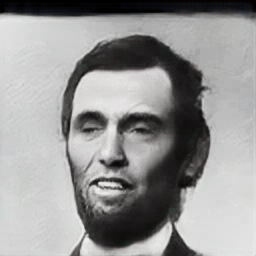} \\
  \includegraphics[width=0.16\linewidth]{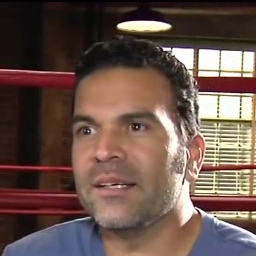} &
  \includegraphics[width=0.16\linewidth]{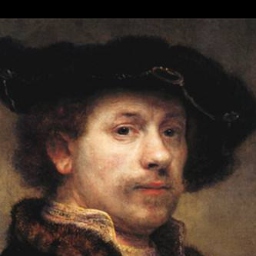} &
  \includegraphics[width=0.16\linewidth]{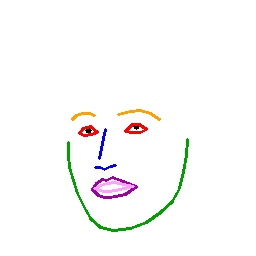} &
  \includegraphics[width=0.16\linewidth]{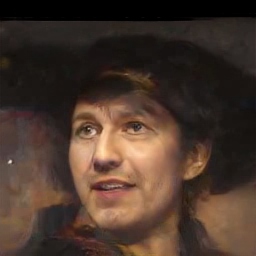} \\
  \includegraphics[width=0.16\linewidth]{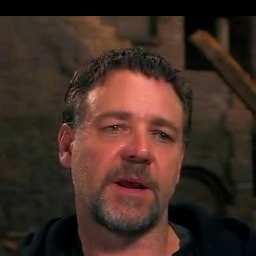} &
  \includegraphics[width=0.16\linewidth]{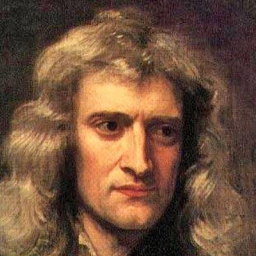} &
  \includegraphics[width=0.16\linewidth]{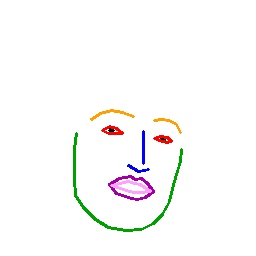} &
  \includegraphics[width=0.16\linewidth]{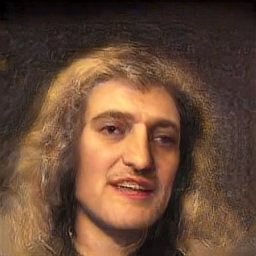} \\
  \includegraphics[width=0.16\linewidth]{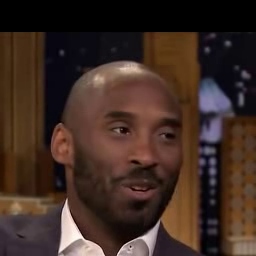} &
  \includegraphics[width=0.16\linewidth]{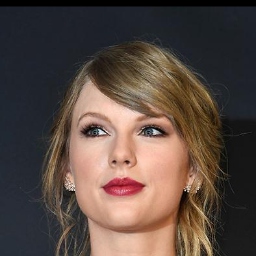} &
  \includegraphics[width=0.16\linewidth]{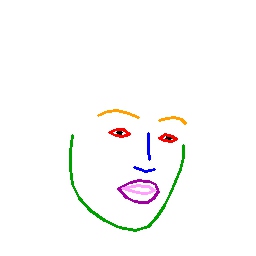} &
  \includegraphics[width=0.16\linewidth]{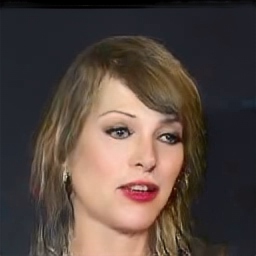} \\
  \includegraphics[width=0.16\linewidth]{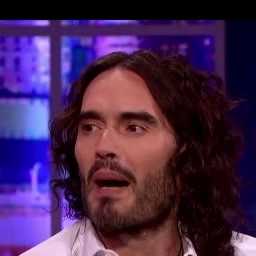} &
  \includegraphics[width=0.16\linewidth]{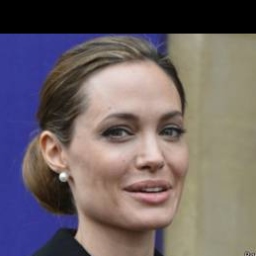} &
  \includegraphics[width=0.16\linewidth]{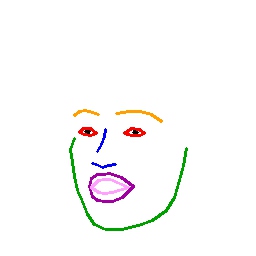} &
  \includegraphics[width=0.16\linewidth]{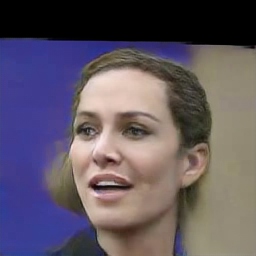} \\
  \includegraphics[width=0.16\linewidth]{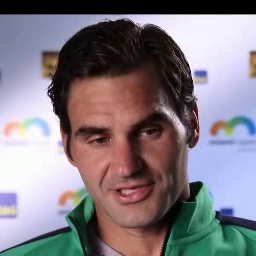} &
  \includegraphics[width=0.16\linewidth]{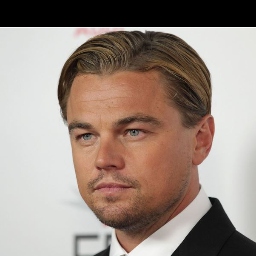} &
  \includegraphics[width=0.16\linewidth]{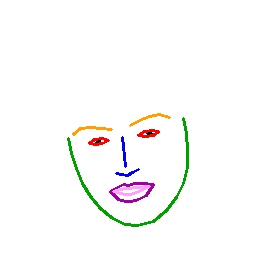} &
  \includegraphics[width=0.16\linewidth]{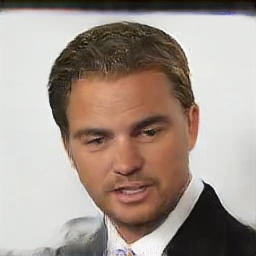} \\
  Source & Target & Landmarks & Result
  \end{tabular}}
\caption{More results.}
	\label{fig:comp7}
\end{figure*}

To demonstrate the capacity and generalization of our method, we test it on in-the-wild face images with the diverse appearance and challenging poses/expressions, including 2D paintings, historical photographs, as well as some celebrity portraits, as shown in Fig.~\ref{fig:comp7}. In the accompanying video, we provide more video examples for reference.
% In the accompanying video, we provide video examples for the comparisons above, as well as video results generated using artworks and historical photographs as the target.

% \subsection{Limitations}
% A major limitation of our method is that we have difficulty in reproducing the hair and background faithfully. The appearance of the target photo is transferred to the source landmark image via a feature dictionary, and we rely on the landmark image to determine in which line the feature at each location should be stored and from which line the feature at each location should be read. Since the facial landmarks contain no structural information about hair or background, it is hard to transfer the appearance of these parts of the target portrait accurately. Please refer to the two failure cases we show in the accompanying video.

\end{document}